%% file: iclr2027_conference.tex
\documentclass{article} % For LaTeX2e
\usepackage{iclr2027_conference,times}

\input{math_commands.tex}

\input{method_names.tex}

\usepackage{hyperref}
\usepackage{url}
\usepackage{algorithm}
\usepackage{algpseudocode}
\usepackage{graphicx}
\input{figure_styles.tex}
\usepackage{framed}
\usepackage{booktabs}
\usepackage{colortbl}
\usepackage{multirow}
\usepackage{dcolumn}
\input{appendix_styles.tex}

\title{\OverallMethod{}: Mitigating Collapse in Iterative Agent Self-Distillation}
\hypersetup{
  pdftitle={ReSAIL: Mitigating Collapse in Iterative Agent Self-Distillation},
  pdfauthor={Shengjie Jin, Hengbo Xu, Zelong Sun, YuJie Guo, Zhiwu Lu}
}

\author{Shengjie Jin\textsuperscript{1} \quad Hengbo Xu\textsuperscript{1} \quad Zelong Sun\textsuperscript{1} \quad YuJie Guo\textsuperscript{1} \quad Zhiwu Lu\textsuperscript{1,*}\\
\textsuperscript{1}Gaoling School of Artificial Intelligence, Renmin University of China, Beijing, China\\
\texttt{\{jinshengjie,luzhiwu\}@ruc.edu.cn}}

\iclrfinalcopy
\begin{document}

\maketitle
\lhead{}
\renewcommand{\headrulewidth}{0pt}
\vspace{-22pt}

\input{sections/00_abstract}
\input{sections/01_introduction}
\input{sections/02_related_work}
\input{sections/03_methodology}
\input{sections/04_experiments}
\input{sections/05_conclusion}

\newpage
\bibliography{iclr2027_conference}
\bibliographystyle{iclr2027_conference}

\input{sections/06_appendix}

\end{document}

%% file: math_commands.tex
\usepackage{amsmath,amsfonts,bm}

\def\eqref#1{equation~\ref{#1}}
\def\1{\bm{1}}

\DeclareMathAlphabet{\mathsfit}{\encodingdefault}{\sfdefault}{m}{sl}
\SetMathAlphabet{\mathsfit}{bold}{\encodingdefault}{\sfdefault}{bx}{n}

%% file: method_names.tex
\newcommand{\OverallMethod}{ReSAIL}
\newcommand{\OverallMethodFull}{\textbf{Re}tentive and \textbf{S}elective \textbf{A}ugmentation for \textbf{I}terative Self-Disti\textbf{l}lation}

\newcommand{\MethodA}{Trajectory-Balanced Selective Distillation}
\newcommand{\MethodASelection}{Sensitivity-Guided Selection}
\newcommand{\MethodABalancing}{Trajectory Loss Balancing}
\newcommand{\MethodB}{Privileged Retention}

\newcommand{\MethodAAbbr}{TBSD}
\newcommand{\MethodASelectionAbbr}{SGS}
\newcommand{\MethodABalancingAbbr}{TLB}
\newcommand{\MethodBAbbr}{PR}

%% file: figure_styles.tex
\usepackage[caption=false]{subfig}
\newlength{\alignedpanelwidth}
\newcommand{\alignedsubfloat}[5]{%
  \begingroup
  \setlength{\alignedpanelwidth}{#2}%
  \captionsetup[subfloat]{margin={#3\alignedpanelwidth,#4\alignedpanelwidth}}%
  \subfloat[#1]{\includegraphics[width=\alignedpanelwidth]{#5}}%
  \endgroup%
}

%% file: appendix_styles.tex
\usepackage{tcolorbox}
\usepackage{listings}
\usepackage{placeins}
\tcbuselibrary{breakable,listings}
\definecolor{eccvblue}{rgb}{0.12,0.49,0.85}

\makeatletter
\newcommand{\appendixtableofcontents}{%
  \section*{Appendix Contents}%
  \begingroup
    \hypersetup{hidelinks,linktoc=all}%
    \parskip=0pt
    \c@tocdepth=2\relax
    \@starttoc{atoc}%
  \endgroup
  \let\appendix@addcontentsline\addcontentsline
  \renewcommand{\addcontentsline}[3]{%
    \appendix@addcontentsline{##1}{##2}{##3}%
    \def\appendix@extension{##1}%
    \def\appendix@toc{toc}%
    \ifx\appendix@extension\appendix@toc
      \addtocontents{atoc}{%
        \protect\contentsline{##2}{##3}{\thepage}{\@currentHref}%
        \protected@file@percent
      }%
    \fi
  }%
}
\makeatother

\newtcblisting{promptlisting}[2][]{
  breakable,
  listing only,
  colback=eccvblue!3,
  colframe=eccvblue!65!black,
  coltitle=black,
  boxrule=0.7pt,
  arc=2mm,
  left=2mm,
  right=2mm,
  top=1mm,
  bottom=1mm,
  title={#2},
  title after break={#2 (continued)},
  fonttitle=\bfseries\fontfamily{ptm}\selectfont,
  colbacktitle=eccvblue!14,
  listing options={
    basicstyle=\fontfamily{ptm}\selectfont\normalsize,
    breaklines=true,
    breakatwhitespace=true,
    breakautoindent=false,
    breakindent=0pt,
    columns=fullflexible,
    keepspaces=true,
    showstringspaces=false,
    literate={—}{{\textemdash}}1 {’}{{\textquoteright}}1
  },
  #1
}

%% file: sections/00_abstract.tex
\begin{abstract}
Iterative self-distillation enables LLM agents to learn from successive deployments, offering a path toward recursive self-improvement (RSI).
Yet our experiments with existing methods reveal a collapse in deployment performance across cycles, while task performance with privileged information (PI) also declines.
We address this collapse by prioritizing informative interaction steps for distillation and preserving PI-conditioned behavior as the student becomes the next teacher.
We introduce \OverallMethodFull{} (\textbf{\OverallMethod{}}), a plug-in augmentation for iterative PI-based self-distillation.
\OverallMethod{} selects interaction steps where PI most strongly changes the teacher's predictions and balances the resulting distillation losses across trajectories.
It also regularizes the student's PI-conditioned output distributions toward those of the frozen teacher at selected and unselected steps to preserve PI-conditioned behavior for supervision in the next cycle.
On ALFWorld and TextCraft, \OverallMethod{} sustains substantial gains across model scales over three cycles, with an average absolute gain of 22.5\% in final-cycle success rates when added to self-distillation baselines.
Sensitivity-guided selection of offline data also improves action prediction accuracy for multimodal GUI agents on AITZ.
These findings provide the first evidence that a more robust learning mechanism can effectively mitigate performance collapse in iterative agent self-distillation over deployment trajectories.
\end{abstract}

%% file: sections/01_introduction.tex
\suppressfloats[t]
\begin{figure}[t]
    \centering
    \alignedsubfloat{Task performance}{0.5\linewidth}{0.16727273}{0.04363636}{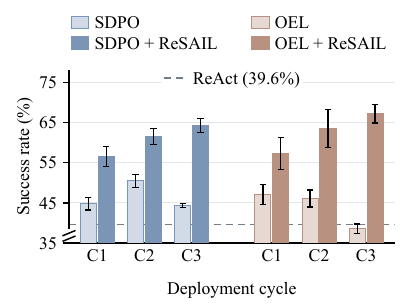}%
    \alignedsubfloat{PI-conditioned task competence}{0.5\linewidth}{0.16727273}{0.04363636}{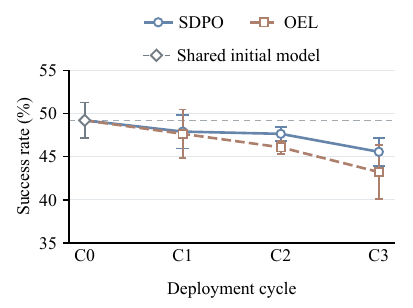}
    \par
    \caption{
\textbf{Iterative self-distillation does not guarantee sustained improvement.}
On ALFWorld OOD with Qwen3-4B:
(a) deployment success rate of SDPO and OEL with and without
\OverallMethod{};
(b) success rate of the original methods conditioned on privileged
information (PI), with the same tasks and their PI held
fixed across cycles. Cycle~0 denotes the shared initial model.
Error bars show the standard deviation of success rates across three decoding seeds.
}
    \label{fig:motivation}
\end{figure}

\section{Introduction}

% 1 
The promise of large language model (LLM) agents extends beyond solving complex tasks to improving through experience~\citep{plaat2025agenticlargelanguagemodels,chen2026recursiveselfimprovementaibounded}.
Through multi-step interactions with their environments, agents generate trajectories of actions, observations, and task outcomes~\citep{yao2023reactsynergizingreasoningacting,wei2026agenticreasoninglargelanguage}.
These trajectories provide feedback for refining future decisions~\citep{shinn2023reflexion}.
Learning from this feedback changes agents' behavior and, in turn, the experience they subsequently generate.
Iterating this feedback loop offers a path toward recursive self-improvement (RSI)~\citep{ye2026online,chen2026recursiveselfimprovementaibounded}.

% 2 
Recent work supports this loop by reusing experience as context or incorporating it into model parameters.
Reflection and retrieval reuse past experience to guide decisions without updating model parameters~\citep{zhao2024expel,zhang2026agenticcontextengineeringevolving}.
Self-distillation instead transfers behavior conditioned on training-only \emph{privileged information} (PI) into model parameters~\citep{shenfeld2026self,hubotter2026reinforcement}.
OEL extends this approach beyond a single training cycle, alternating experience collection and offline self-distillation across successive deployments~\citep{ye2026online}.
Yet repeating this cycle does not guarantee continued improvement~\citep{chen2026rethinking}.
Our experiments with SDPO and OEL show that performance can collapse over successive cycles (Figure~\ref{fig:motivation}(a)).

% 3
To investigate what may limit continued improvement, we ask two questions about the supervision used in iterative self-distillation.
\textbf{(Q1) Which interaction steps should be prioritized for distillation?}
We hypothesize that interaction steps where PI most strongly changes the teacher's predictions offer particularly informative supervision for transferring PI-conditioned behavior.
We quantify this influence using \emph{privileged-information sensitivity} (PI sensitivity), defined as the Jensen--Shannon divergence between the teacher's output distributions with and without PI at the same interaction step.
\textbf{(Q2) What happens when the student becomes the next teacher?}
Each updated model serves two roles: it acts without PI during deployment and provides PI-conditioned supervision in the next cycle.
The base objective (Equation~\ref{eq:base-objective}) trains the student only without PI and does not explicitly preserve its PI-conditioned behavior.
With evaluation tasks and PI held fixed, both SDPO and OEL show declining mean PI-conditioned task success over successive cycles (Figure~\ref{fig:motivation}(b)).
These considerations motivate selective distillation within each cycle and preservation of PI-conditioned competence as the updated student becomes the next teacher.

% 4
Guided by these considerations, we propose \OverallMethodFull{} (\textbf{\OverallMethod{}}), a plug-in augmentation for iterative PI-based self-distillation.
The approach combines two complementary modules: \textbf{\MethodA{}} (\MethodAAbbr{}) for selective distillation within each cycle and \textbf{\MethodB{}} (\MethodBAbbr{}) for retaining PI-conditioned behavior across cycles.
\MethodAAbbr{} uses \MethodASelection{} (\MethodASelectionAbbr{}) to prioritize interaction steps with the highest teacher PI sensitivity and balances the selected losses across trajectories.
\MethodBAbbr{} regularizes the student's privileged-view output distributions toward those of the frozen teacher at both selected and unselected interaction steps.
This retention objective targets the PI-conditioned behavior used when the updated student becomes the teacher in the next cycle.

% 5
Our contributions are summarized as follows:
(1) We provide the first evidence that performance collapse in iterative agent self-distillation can be effectively mitigated through a more robust learning mechanism.
(2) We propose \OverallMethod{}, a plug-in augmentation that combines trajectory-balanced selective distillation with privileged retention to improve learning within each cycle and preserve PI-conditioned behavior across cycles.
(3) Experiments on ALFWorld and TextCraft demonstrate substantial and sustained performance gains across model scales over three deployment cycles.
\OverallMethod{} improves final-cycle success rates by an average of 22.5 percentage points over the corresponding SDPO and OEL baselines.
We further demonstrate that \MethodASelection{} improves action prediction accuracy for multimodal GUI agents on AITZ.

%% file: sections/02_related_work.tex
\section{Related Work}

\paragraph{Learning and Adaptation in LLM Agents}
LLM agents have been widely explored as autonomous systems that combine reasoning, tool use, and interaction with environments to accomplish tasks~\citep{yao2023reactsynergizingreasoningacting,plaat2025agenticlargelanguagemodels,wei2026agenticreasoninglargelanguage}.
Context-based approaches enable agents to learn from past interactions by reusing reflections and extracted insights to guide subsequent decisions, without updating model parameters~\citep{shinn2023reflexion,zhao2024expel,zhang2026agenticcontextengineeringevolving,ouyang2026reasoningbankscalingagentselfevolving}.
Other approaches instead update model parameters through supervised fine-tuning on interaction trajectories~\citep{dou2024re,yuan2025agentrtraininglanguagemodel} or reinforcement learning from environmental rewards~\citep{NEURIPS2025_420c9f77,wang2025ragenunderstandingselfevolutionllm,ICLR2026_1571ce1f}.
Recent work further explores how to reuse accumulated interaction trajectories to support subsequent agent learning and adaptation~\citep{ye2026online,wu2026terminaluniverseturningagenttrajectories}.
Within this line of work, we investigate how agent self-distillation can remain effective across successive deployment cycles.

\paragraph{Self-Distillation with Privileged Information}
Self-distillation with privileged information uses a model's behavior conditioned on additional context to supervise its own learning, without requiring that context at deployment~\citep{shenfeld2026self,hubotter2026reinforcement}.
Here, privileged information (PI) refers to auxiliary information used during training but omitted from the deployed policy's inputs~\citep{penaloza2026privilegedinformationdistillationlanguage}.
PI can include reference solutions or demonstrations~\citep{shenfeld2026self}, hindsight feedback~\citep{hubotter2026reinforcement}, and knowledge extracted from interaction trajectories~\citep{ye2026online,wang2026skillsdskillconditionedselfdistillationmultiturn}.
In each case, the quality of self-distillation supervision depends on how effectively the model can use the supplied PI.
We prioritize self-distillation at interaction steps where PI most strongly changes the teacher's predictions.

\paragraph{Iterative Self-Distillation}
OEL alternates experience collection during deployment with offline self-distillation of knowledge extracted from that experience~\citep{ye2026online}.
However, repeated self-distillation can degrade agent performance across successive cycles~\citep{chen2026rethinking}.
To mitigate this degradation, \citet{chen2026rethinking} shift to off-policy context-distillation with a forward-KL objective on successful trajectories generated by PI-conditioned teachers.
Collecting these teacher trajectories requires additional environment interaction, making this approach inapplicable when the deployment environment cannot be revisited during training.
In contrast, \OverallMethod{} addresses this practical and more restrictive setting by strengthening self-distillation on already collected deployment trajectories.
It combines selective distillation within each cycle with a retention objective that targets the PI-conditioned behavior used when the updated student becomes the next teacher.

%% file: sections/03_methodology.tex
\begin{figure}[t]
    \centering
    % \vspace{-5pt}
    \includegraphics[width=\linewidth]{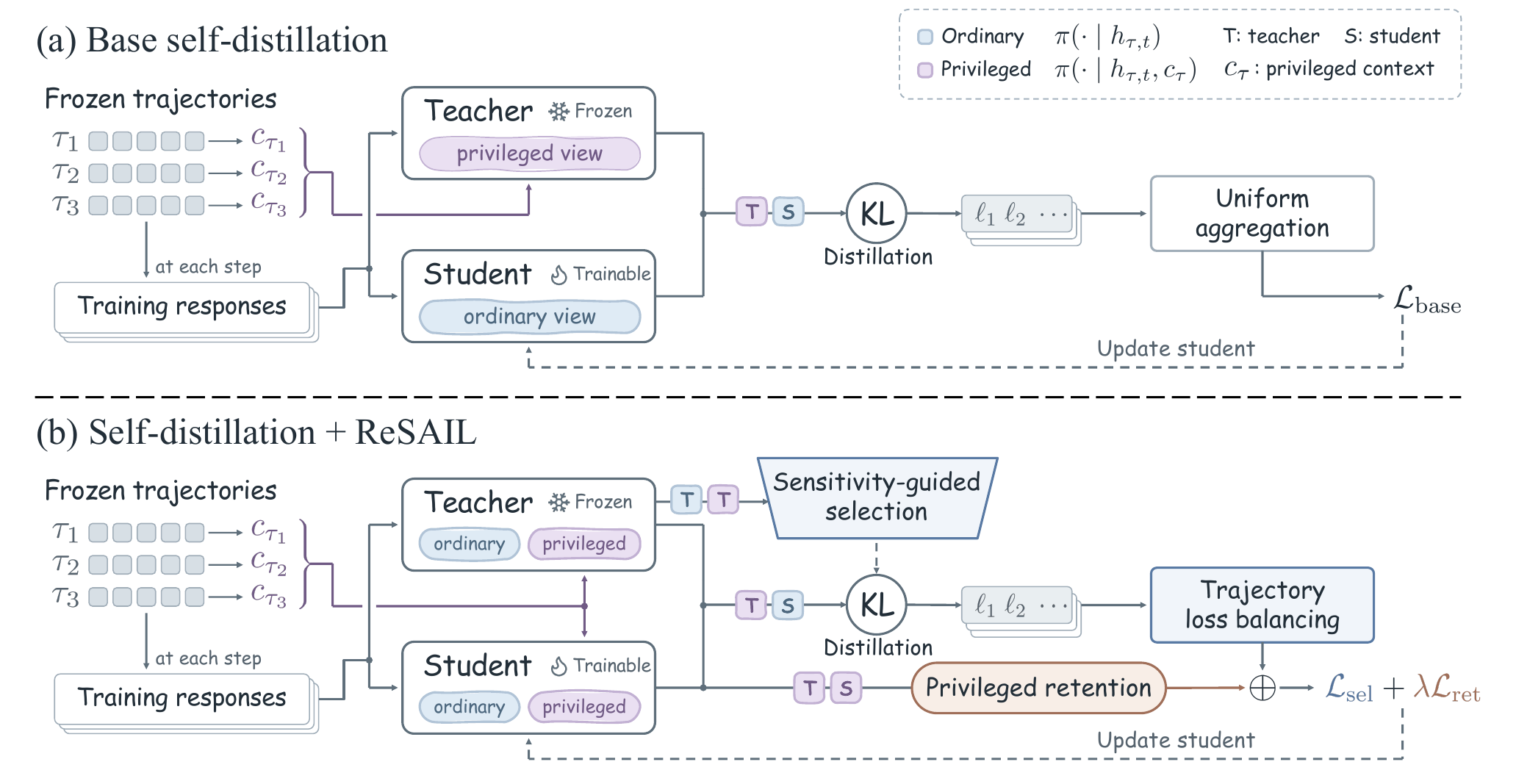}
    \vspace{-15pt}
    \caption{\textbf{Overview of \OverallMethod{}}.
    (a) Base self-distillation matches the ordinary-view student to the privileged-view teacher, uniformly averaging losses over all interaction steps used for training.
    (b) \OverallMethod{} uses \textbf{\MethodASelection{}} (\MethodASelectionAbbr{}) to select the most PI-sensitive steps and \textbf{\MethodABalancing{}} (\MethodABalancingAbbr{}) to balance the selected losses across trajectories.
    \textbf{\MethodB{}} (\MethodBAbbr{}) regularizes the student's privileged-view predictions toward those of the teacher at both selected and unselected interaction steps within each training batch.}
    \label{fig:method_overview}
    \vspace{-2pt}
\end{figure}

\section{Methodology}

In this section, we present \textbf{\OverallMethod{}}, a plug-in augmentation for iterative self-distillation with privileged information (PI).
We first introduce the offline learning setup and the base self-distillation objective (Section~\ref{sec:method_setup}).
We then describe \textbf{\MethodA{}} (Section~\ref{sec:selective_distillation}) and \textbf{\MethodB{}} (Section~\ref{sec:privileged_retention}), followed by the joint objective and optimization procedure (Section~\ref{sec:joint_optimization}).
Figure~\ref{fig:method_overview} provides an overview of \OverallMethod{}.

\subsection{Setup and Base Self-Distillation}
\label{sec:method_setup}

In cycle \(r\), a starting policy \(\pi_r\) is paired with a trajectory dataset
\(\mathcal D_r\), collected by \(\pi_r\) or supplied by an external offline source.
Training uses this fixed dataset without further environment interaction.
Each trajectory is represented as
\[
    \tau=\bigl(h_{\tau,t},\tilde y_{\tau,t}\bigr)_{t=1}^{|\tau|},
    \qquad \tau\in\mathcal D_r,
\]
where \(h_{\tau,t}\) is the interaction history at interaction step \(t\) and \(\tilde y_{\tau,t}\) is the logged response.

For each trajectory, the base method supplies PI \(c_\tau\), such as the complete trajectory or an extracted experience.
A policy \(\pi\) has two views that share parameters:
\[
    \underbrace{\pi(\cdot\mid h_{\tau,t})}_{\text{ordinary view}}
    \qquad\text{and}\qquad
    \underbrace{\pi(\cdot\mid h_{\tau,t},c_\tau)}_{\text{privileged view}}.
\]
For offline training, we initialize the student \(\pi^{\mathrm S}\) and a frozen teacher \(\pi^{\mathrm T}\) from \(\pi_r\).
The updated student defines \(\pi_{r+1}\), acts without PI at deployment,
and initializes both models in the next cycle.

During offline training, the base method supplies a training response at each logged step,
\[
    z_{\tau,t}\sim\mu(\cdot\mid h_{\tau,t},\tau).
\]
In our OEL~\citep{ye2026online} integration, \(z_{\tau,t}\) is freshly sampled from the ordinary-view student at each update.
In our SDPO-offline~\citep{hubotter2026reinforcement} integration, it is generated once by the cycle-start policy and reused throughout the cycle.
At each logged step \((\tau,t)\), all distributional comparisons use the same prefixes \(z_{\tau,t,<k}\) of the training response \(z_{\tau,t}\).

For a nonempty trajectory batch \(\mathcal B\subseteq\mathcal D_r\), let \(\mathcal U\subseteq\{(\tau,t):\tau\in\mathcal B,\ 1\leq t\leq|\tau|\}\) denote the set of interaction steps used by the base training objective before sensitivity-guided selection.
At each step in \(\mathcal U\), the ordinary-view student matches the privileged-view teacher through token-averaged student-to-teacher KL divergence:
\begin{equation}
    \label{eq:distillation-loss}
    \ell_{\mathrm{dist}}(\tau,t)
    =\frac{1}{|z_{\tau,t}|}
        \sum_{k=1}^{|z_{\tau,t}|}
        D_{\mathrm{KL}}\!\left(
            \pi^{\mathrm S}(\cdot\mid h_{\tau,t},z_{\tau,t,<k})
            \,\middle\|\,
            \pi^{\mathrm T}(\cdot\mid h_{\tau,t},c_\tau,z_{\tau,t,<k})
        \right).
\end{equation}
Averaging these losses over steps in \(\mathcal U\) gives the base objective:
\begin{equation}
    \label{eq:base-objective}
    \mathcal L_{\mathrm{base}}
    =\frac{1}{|\mathcal U|}
    \sum_{(\tau,t)\in\mathcal U}
    \ell_{\mathrm{dist}}(\tau,t),
    \qquad \mathcal U\neq\emptyset.
\end{equation}
The influence of PI on the teacher's predictions varies across steps.
We therefore concentrate distillation on steps where PI induces
the largest prediction changes.

\subsection{\MethodA{} (\MethodAAbbr{})}
\label{sec:selective_distillation}

\paragraph{\MethodASelection{} (\MethodASelectionAbbr{}).}
We quantify this prediction shift using PI sensitivity.
For any policy \(\pi\), history \(h\), PI \(c\), and response \(z\), we define \emph{PI sensitivity} as the average divergence between its ordinary and privileged views:
\begin{equation}
    \label{eq:privileged-sensitivity}
    s_\pi(h,c;z)
    =\frac{1}{|z|}\sum_{k=1}^{|z|}
    \operatorname{JSD}\!\left(
        \pi(\cdot\mid h,z_{<k}),
        \pi(\cdot\mid h,c,z_{<k})
    \right),
\end{equation}
where \(\operatorname{JSD}\) denotes the Jensen--Shannon divergence.
We score each step in \(\mathcal U\) using the frozen teacher's PI sensitivity,
$s(\tau,t):=s_{\pi^{\mathrm T}}(h_{\tau,t},c_\tau;z_{\tau,t})$.
Because both views use the same model parameters and response prefixes, the score measures the prediction change induced by PI.
For a selection ratio $\rho\in(0,1]$, we retain the highest-scoring steps across the entire batch:
\begin{equation}
    \label{eq:method-a-selection}
    K_\rho=\left\lceil\rho|\mathcal U|\right\rceil,
    \qquad
    \mathcal C_\rho=\operatorname{TopK}(\mathcal U;s,K_\rho).
\end{equation}

\paragraph{\MethodABalancing{} (\MethodABalancingAbbr{}).}
A uniform average over selected steps gives greater total weight to trajectories with more selected steps.
We therefore average the selected losses within each trajectory and normalize their sum by the full trajectory-batch size. Let $\mathcal C_\rho(\tau)=\{t:(\tau,t)\in\mathcal C_\rho\}$ and $m_\tau=|\mathcal C_\rho(\tau)|$.
The selected distillation objective is
\begin{equation}
    \label{eq:selected-distillation-objective}
    \mathcal L_{\mathrm{sel}}
    =\frac{1}{|\mathcal B|}
    \sum_{\substack{\tau\in\mathcal B\\m_\tau>0}}
    \frac{1}{m_\tau}
    \sum_{t\in\mathcal C_\rho(\tau)}
    \ell_{\mathrm{dist}}(\tau,t).
\end{equation}
Each trajectory with selected steps receives total loss coefficient \(1/|\mathcal B|\); the others contribute zero.

\subsection{\MethodB{} (\MethodBAbbr{})}
\label{sec:privileged_retention}

The updated student also serves as the next cycle's PI-conditioned teacher.
Because its two views share parameters, ordinary-view distillation can alter the PI-conditioned behavior used for future supervision.
To retain this behavior for the next cycle, \MethodBAbbr{} matches the student's privileged-view output distributions to those of the current frozen teacher:
\begin{equation}
    \label{eq:privileged-retention}
    \ell_{\mathrm{ret}}(\tau,t)
    =\frac{1}{|z_{\tau,t}|}
        \sum_{k=1}^{|z_{\tau,t}|}
        D_{\mathrm{KL}}\!\left(
            \pi^{\mathrm S}(\cdot\mid h_{\tau,t},c_\tau,z_{\tau,t,<k})
            \,\middle\|\,
            \pi^{\mathrm T}(\cdot\mid h_{\tau,t},c_\tau,z_{\tau,t,<k})
        \right).
\end{equation}
Retention covers all steps in \(\mathcal U\), including those not selected for distillation, using the same trajectory-level reduction as \MethodABalancingAbbr{}.
With \(\mathcal U(\tau)=\{t:(\tau,t)\in\mathcal U\}\),
\begin{equation}
    \label{eq:retention-objective}
    \mathcal L_{\mathrm{ret}}
    =\frac{1}{|\mathcal B|}
    \sum_{\substack{\tau\in\mathcal B\\|\mathcal U(\tau)|>0}}
    \frac{1}{|\mathcal U(\tau)|}
    \sum_{t\in\mathcal U(\tau)}
    \ell_{\mathrm{ret}}(\tau,t).
\end{equation}

\subsection{Joint Objective and Optimization}
\label{sec:joint_optimization}

We combine selective distillation and privileged retention as
\begin{equation}
    \label{eq:joint-objective}
    \mathcal L
    =\mathcal L_{\mathrm{sel}}+\lambda\mathcal L_{\mathrm{ret}},
    \qquad \lambda\geq0,
\end{equation}
where $\lambda$ controls the retention strength.
At each update, we recompute the teacher's sensitivity scores and selected set, holding both fixed during backpropagation.
Algorithm~\ref{alg:method_training} in Appendix~\ref{app:implementation} details one internalization cycle, including tie-breaking and empty-set handling.

%% file: sections/04_experiments.tex
\section{Experiments}

\subsection{Experimental Setup}

\paragraph{Benchmarks and Models.}
We evaluate \OverallMethod{} on ALFWorld~\citep{ALFWorld20} for embodied
household task planning and TextCraft~\citep{prasad-etal-2024-adapt} for
compositional crafting. AITZ~\citep{zhang-etal-2024-android} tests
\MethodASelectionAbbr{} as an offline data filter for Android GUI navigation.
We use Qwen3-4B and Qwen3-8B~\citep{qwen3} with thinking mode disabled for the text-based benchmarks, and Qwen3-VL-4B-Instruct~\citep{Qwen3-VL} for AITZ.

\paragraph{Baselines.}
On the text benchmarks, we compare against ReAct~\citep{yao2023reactsynergizingreasoningacting},
RFT~\citep{yuan2023scalingrelationshiplearningmathematical},
offline GRPO~\citep{shao2024deepseekmathpushinglimitsmathematical},
EPD~\citep{gou2026sample},
offline SDPO~\citep{hubotter2026reinforcement}, and OEL~\citep{ye2026online}.
All text-task PI methods use experience summaries from their own complete trajectories.
We evaluate \OverallMethod{} as a plug-in to SDPO and OEL on text tasks.
On AITZ, we compare OEL with OEL + \MethodASelectionAbbr{}, using
\MethodASelectionAbbr{} only to filter training data before optimization.
Implementation details are provided in Appendix~\ref{app:baselines}.

\paragraph{Training Protocol.}
Text agents undergo three cycles of 30 updates, each using fresh
trajectories and initializing from their own preceding final checkpoint.
Update counts and trajectory batch sizes are matched within each text setting.
AITZ uses one cycle of 100 updates with 16 source trajectories per update
for both methods. OEL + \MethodASelectionAbbr{} retains a fixed global
top 80\% of interaction steps before optimization. Data preparation and implementation details are provided in
Appendices~\ref{app:experimental_settings} and~\ref{app:implementation}.

\paragraph{Evaluation.}
We evaluate final checkpoints after each text-task cycle using task success
rate on ALFWorld (ID/OOD) and TextCraft. ReAct is the unchanged initial
policy for text tasks. These final results report means and standard
deviations (SD) across three decoding seeds. On AITZ, we monitor offline
step-level accuracy every ten updates on a fixed test subset, excluding
reference \texttt{stop} steps and following the official action-matching rules.
Detailed evaluation settings are provided in Appendix~\ref{app:evaluation}.
For the retention analysis, we evaluate all checkpoints on the same 128 ALFWorld OOD tasks with matched initial states. The task-specific PI consists of summaries generated by the initial model from separate reference episodes and is held fixed across evaluations (Appendix~\ref{app:dual_view_evaluation}).

\subsection{Main Results}

\begin{table}[t]

    \centering

    \caption{\textbf{Main results across three
deployment cycles.}
Task success rates (\%) without PI, reported as mean (SD)
over three decoding seeds. SD denotes standard deviation.
Shaded rows add \OverallMethod{} to the preceding baseline;
bold marks the highest mean in each column within each model size.
ReAct is the unchanged initial policy, shown once under Cycle~1 as a fixed reference across cycles.}

    \label{tab:main_results}
    \vspace{5pt}

    \small

    \setlength{\tabcolsep}{0pt}

    \renewcommand{\arraystretch}{1.45}

    \newcommand{\resultcell}[2]{%
        \begingroup
        \renewcommand{\arraystretch}{1}%
        \begin{tabular}[t]{@{}D{.}{.}{2.1}@{\hspace{1pt}}l@{}}
            #1 & {\scriptsize(#2)}
        \end{tabular}%
        \endgroup}

    \newcommand{\bestresultcell}[2]{%
        \begingroup\bfseries\boldmath
        \resultcell{#1}{#2}%
        \endgroup}

    \dimen0=\dimexpr(\textwidth-0.18\textwidth-10pt)/9\relax

    \resizebox{\textwidth}{!}{%

    \begin{tabular}{@{}
        >{\centering\arraybackslash}p{0.15\textwidth}
        >{\raggedright\arraybackslash}p{0.13\textwidth}
        *{3}{>{\centering\arraybackslash}p{\dimen0}}p{5pt}
        *{3}{>{\centering\arraybackslash}p{\dimen0}}p{5pt}
        *{3}{>{\centering\arraybackslash}p{\dimen0}}@{}}

        \toprule

        \multirow{2}{*}{\textbf{Model}} & \multirow{2}{*}{\textbf{Method}}
        & \multicolumn{3}{c}{\textbf{ALFWorld (ID)}} &
        & \multicolumn{3}{c}{\textbf{ALFWorld (OOD)}} &
        & \multicolumn{3}{c}{\textbf{TextCraft}} \\

        \cmidrule(lr){3-5}
        \cmidrule(lr){7-9}
        \cmidrule(lr){11-13}

        & & \multicolumn{1}{c}{Cycle~1}
          & \multicolumn{1}{c}{Cycle~2}
          & \multicolumn{1}{c}{Cycle~3}
        & & \multicolumn{1}{c}{Cycle~1}
          & \multicolumn{1}{c}{Cycle~2}
          & \multicolumn{1}{c}{Cycle~3}
        & & \multicolumn{1}{c}{Cycle~1}
          & \multicolumn{1}{c}{Cycle~2}
          & \multicolumn{1}{c}{Cycle~3} \\

        \midrule

        \multirow{8}{*}{Qwen3-4B}

        & ReAct
        & \resultcell{40.1}{2.3} & -- & --
        & & \resultcell{39.6}{1.8} & -- & --
        & & \resultcell{61.3}{1.5} & -- & -- \\

        & RFT
        & \resultcell{43.2}{2.7} & \resultcell{44.8}{4.4} & \resultcell{41.9}{6.6}
        & & \resultcell{39.3}{4.6} & \resultcell{42.2}{1.4} & \resultcell{43.2}{3.9}
        & & \resultcell{59.7}{0.6} & \resultcell{62.3}{5.5} & \resultcell{60.7}{2.5} \\

        & GRPO
        & \resultcell{39.1}{2.1} & \resultcell{36.5}{1.2} & \resultcell{37.5}{2.3}
        & & \resultcell{45.1}{1.2} & \resultcell{42.2}{1.6} & \resultcell{50.3}{2.0}
        & & \resultcell{60.3}{1.2} & \resultcell{58.7}{2.3} & \resultcell{56.7}{2.3} \\

        & EPD
        & \resultcell{45.6}{1.2} & \resultcell{36.7}{3.4} & \resultcell{22.1}{0.5}
        & & \resultcell{43.2}{2.7} & \resultcell{31.5}{3.2} & \resultcell{18.2}{3.5}
        & & \resultcell{63.3}{0.6} & \resultcell{62.0}{2.6} & \resultcell{62.3}{3.5} \\

        & SDPO
        & \resultcell{52.3}{1.6} & \resultcell{53.4}{2.0} & \resultcell{58.9}{1.6}
        & & \resultcell{44.8}{1.6} & \resultcell{50.5}{1.6} & \resultcell{44.3}{0.5}
        & & \resultcell{62.7}{2.3} & \resultcell{64.7}{0.6} & \resultcell{60.3}{4.5} \\

        \rowcolor{black!5}
        \cellcolor{white}
        & \hspace{0.6em}\textbf{+~\OverallMethod{}}
        & \bestresultcell{66.4}{2.1} & \bestresultcell{72.1}{1.6} & \resultcell{72.4}{0.9}
        & & \resultcell{56.5}{2.5} & \resultcell{61.5}{2.0} & \resultcell{64.3}{1.8}
        & & \bestresultcell{64.7}{1.2} & \bestresultcell{66.3}{2.9} & \bestresultcell{67.3}{1.2} \\

        & OEL
        & \resultcell{54.9}{4.7} & \resultcell{56.5}{2.0} & \resultcell{47.4}{1.2}
        & & \resultcell{47.1}{2.5} & \resultcell{46.1}{2.1} & \resultcell{38.5}{1.2}
        & & \resultcell{62.0}{3.6} & \resultcell{54.7}{1.5} & \resultcell{43.7}{1.5} \\

        \rowcolor{black!5}
        \cellcolor{white}
        & \hspace{0.6em}\textbf{+~\OverallMethod{}}
        & \resultcell{62.2}{2.0} & \resultcell{70.1}{3.2} & \bestresultcell{74.2}{1.4}
        & & \bestresultcell{57.3}{4.0} & \bestresultcell{63.5}{4.7} & \bestresultcell{67.2}{2.3}
        & & \resultcell{63.3}{1.5} & \resultcell{64.0}{3.6} & \resultcell{64.0}{2.6} \\

        \midrule

        \multirow{8}{*}{Qwen3-8B}

        & ReAct
        & \resultcell{49.5}{2.4} & -- & --
        & & \resultcell{57.0}{3.9} & -- & --
        & & \resultcell{64.0}{1.0} & -- & -- \\

        & RFT
        & \resultcell{50.0}{2.8} & \resultcell{52.1}{2.5} & \resultcell{52.1}{2.7}
        & & \resultcell{60.4}{1.2} & \resultcell{61.2}{1.8} & \resultcell{63.8}{5.3}
        & & \resultcell{66.0}{4.4} & \resultcell{62.7}{2.5} & \resultcell{58.7}{2.1} \\

        & GRPO
        & \resultcell{53.1}{4.9} & \resultcell{47.4}{1.2} & \resultcell{48.2}{2.5}
        & & \resultcell{62.8}{5.0} & \resultcell{65.4}{3.5} & \resultcell{62.2}{1.2}
        & & \resultcell{57.0}{2.6} & \resultcell{64.3}{2.1} & \resultcell{59.3}{2.1} \\

        & EPD
        & \resultcell{68.0}{2.3} & \resultcell{67.2}{2.8} & \resultcell{53.4}{2.7}
        & & \resultcell{62.0}{1.6} & \resultcell{56.0}{1.2} & \resultcell{46.6}{3.9}
        & & \resultcell{65.7}{2.3} & \resultcell{65.0}{2.6} & \resultcell{64.3}{0.6} \\

        & SDPO
        & \resultcell{71.4}{1.6} & \resultcell{72.1}{1.2} & \resultcell{61.2}{3.9}
        & & \resultcell{66.1}{2.7} & \resultcell{64.1}{4.1} & \resultcell{49.2}{2.8}
        & & \resultcell{63.0}{1.0} & \resultcell{62.3}{2.3} & \resultcell{51.0}{2.0} \\

        \rowcolor{black!5}
        \cellcolor{white}
        & \hspace{0.6em}\textbf{+~\OverallMethod{}}
        & \bestresultcell{75.5}{1.8} & \bestresultcell{80.5}{1.4} & \resultcell{82.0}{1.4}
        & & \bestresultcell{72.1}{5.2} & \resultcell{73.2}{0.9} & \resultcell{75.3}{1.2}
        & & \resultcell{65.7}{2.3} & \resultcell{70.7}{3.1} & \resultcell{73.3}{1.5} \\

        & OEL
        & \resultcell{72.4}{0.5} & \resultcell{73.4}{3.9} & \resultcell{58.3}{0.9}
        & & \resultcell{66.7}{2.7} & \resultcell{61.7}{5.5} & \resultcell{41.4}{1.4}
        & & \resultcell{63.0}{2.6} & \resultcell{61.3}{1.5} & \resultcell{51.3}{0.6} \\

        \rowcolor{black!5}
        \cellcolor{white}
        & \hspace{0.6em}\textbf{+~\OverallMethod{}}
        & \resultcell{73.7}{2.7} & \resultcell{79.2}{1.2} & \bestresultcell{82.8}{1.6}
        & & \resultcell{70.8}{3.5} & \bestresultcell{73.4}{2.3} & \bestresultcell{78.4}{1.2}
        & & \bestresultcell{69.0}{2.6} & \bestresultcell{72.3}{1.2} & \bestresultcell{73.7}{1.5} \\

        \bottomrule

    \end{tabular}%
    }

\end{table}

\paragraph{Higher Final-Cycle Performance.}
After three deployment cycles, both SDPO+\OverallMethod{} and
OEL+\OverallMethod{} outperform all unaugmented baselines across
ALFWorld ID/OOD and TextCraft at both model sizes
(Table~\ref{tab:main_results}).
Relative to their respective parent methods, \OverallMethod{}
improves final-cycle success by 7.0--26.1 percentage points for SDPO
and 20.3--37.0 points for OEL.
The gains also hold against the strongest alternative baseline.
On ALFWorld OOD with Qwen3-8B, SDPO+\OverallMethod{} and
OEL+\OverallMethod{} achieve 75.3\% and 78.4\% success, respectively.
The corresponding success rates are 63.8\% for RFT, 62.2\% for GRPO,
and 46.6\% for EPD.

\paragraph{Sustained Gains Across Deployment Cycles.}
Both augmented methods outperform their respective parent methods
at every reported cycle and maintain non-decreasing mean success
across all three cycles in every evaluated setting.
By contrast, OEL finishes below its first-cycle performance in all
six settings, and SDPO does so in five.
The benefit of \OverallMethod{} therefore extends beyond an
initial improvement.
Averaged equally across the six model--benchmark settings, the advantage over the corresponding parent method grows from 6.8 to 18.3 percentage points for SDPO and from 5.0 to 26.6 points for OEL between the first and third deployment cycles.

\newpage
\subsection{Ablation Studies}
\label{sec:ablation_studies}

\paragraph{Selection and Retention Contribute at Different Timescales.}
The component ablation shows complementary roles for selective
distillation and privileged-view retention
(Table~\ref{tab:component_ablation}).
\MethodASelectionAbbr{} alone achieves the strongest first-cycle
performance, improving over OEL by 11.5 and 14.4 percentage points
on ID and OOD, respectively.
Adding \MethodABalancingAbbr{} improves third-cycle success over
selection alone by 6.0 and 7.8 points, but the resulting
\MethodAAbbr{} configuration still finishes below its first-cycle
performance.
Adding \MethodBAbbr{} changes this cross-cycle trajectory:
the full method achieves the highest success in Cycles~2 and~3,
exceeding \MethodAAbbr{} by 17.7 and 21.6 points at Cycle~3.
The full method also outperforms \MethodBAbbr{} alone by 9.9 and
13.3 points, showing that selective distillation remains valuable
when training already includes a privileged-view retention objective.

\begin{table}[t]
    \centering
    \caption{\textbf{Component ablations} with OEL and Qwen3-4B on ALFWorld.
$\checkmark$ and $\times$ indicate enabled and disabled components;
shading marks the full method.
Entries report success rates (\%) as mean (SD) over three
decoding seeds. Bold marks the highest mean in each column.
    \MethodASelectionAbbr{}: \MethodASelection{};
    \MethodABalancingAbbr{}: \MethodABalancing{};
    \MethodBAbbr{}: \MethodB{}.}
    \label{tab:component_ablation}
    \vspace{5pt}
    \small
    \setlength{\tabcolsep}{0pt}
    \renewcommand{\arraystretch}{1.3}
    \newcommand{\ablationresult}[2]{%
        \begingroup\renewcommand{\arraystretch}{1}%
        \begin{tabular}[t]{@{}D{.}{.}{2.1}@{\hspace{1pt}}l@{}}
            #1 & {\scriptsize(#2)}
        \end{tabular}%
        \endgroup}
    \newcommand{\bestablationresult}[2]{%
        \begingroup\bfseries\boldmath
        \ablationresult{#1}{#2}%
        \endgroup}
    \begin{tabular}{@{}
        *{3}{>{\centering\arraybackslash}p{0.075\textwidth}}
        p{0.035\textwidth}
        *{3}{>{\centering\arraybackslash}p{0.12\textwidth}}
        p{0.02\textwidth}
        *{3}{>{\centering\arraybackslash}p{0.12\textwidth}}@{}}
        \toprule
        \multicolumn{3}{c}{\textbf{Components}} &
        & \multicolumn{3}{c}{\textbf{ALFWorld (ID)}} &
        & \multicolumn{3}{c}{\textbf{ALFWorld (OOD)}} \\
        \cmidrule(lr){1-3}\cmidrule(lr){5-7}\cmidrule(lr){9-11}
        \MethodASelectionAbbr{} & \MethodABalancingAbbr{} & \MethodBAbbr{} &
        & Cycle 1 & Cycle 2 & Cycle 3 &
        & Cycle 1 & Cycle 2 & Cycle 3 \\
        \midrule
        \(\textcolor{black!40}{\times}\) & \(\textcolor{black!40}{\times}\) & \(\textcolor{black!40}{\times}\) &  & \ablationresult{54.9}{4.7} & \ablationresult{56.5}{2.0} & \ablationresult{47.4}{1.2} &  & \ablationresult{47.1}{2.5} & \ablationresult{46.1}{2.1} & \ablationresult{38.5}{1.2} \\
        \(\checkmark\) & \(\textcolor{black!40}{\times}\) & \(\textcolor{black!40}{\times}\) &  & \bestablationresult{66.4}{4.8} & \ablationresult{69.0}{4.6} & \ablationresult{50.5}{3.2} &  & \bestablationresult{61.5}{5.9} & \ablationresult{56.5}{2.5} & \ablationresult{37.8}{2.5} \\
        \(\textcolor{black!40}{\times}\) & \(\checkmark\) & \(\textcolor{black!40}{\times}\) &  & \ablationresult{56.5}{1.6} & \ablationresult{60.2}{2.1} & \ablationresult{59.9}{2.4} &  & \ablationresult{51.3}{1.6} & \ablationresult{50.8}{2.7} & \ablationresult{50.8}{3.1} \\
        \(\textcolor{black!40}{\times}\) & \(\textcolor{black!40}{\times}\) & \(\checkmark\) &  & \ablationresult{52.9}{4.4} & \ablationresult{55.2}{0.9} & \ablationresult{64.3}{1.6} &  & \ablationresult{49.0}{2.0} & \ablationresult{49.7}{1.6} & \ablationresult{53.9}{2.1} \\
        \(\checkmark\) & \(\checkmark\) & \(\textcolor{black!40}{\times}\) &  & \ablationresult{65.9}{3.0} & \ablationresult{68.5}{0.5} & \ablationresult{56.5}{3.0} &  & \ablationresult{60.4}{3.0} & \ablationresult{58.9}{1.6} & \ablationresult{45.6}{1.6} \\
        \rowcolor{black!5}
        \(\checkmark\) & \(\checkmark\) & \(\checkmark\) &  & \ablationresult{62.2}{2.0} & \bestablationresult{70.1}{3.2} & \bestablationresult{74.2}{1.4} &  & \ablationresult{57.3}{4.0} & \bestablationresult{63.5}{4.7} & \bestablationresult{67.2}{2.3} \\
        \bottomrule
    \end{tabular}
\end{table}

\begin{table}[t]
    \centering
    \caption{\textbf{Selection strategy and retention view.}
OEL with Qwen3-4B on ALFWorld.
(a) Cycle~1 selection at $\rho=0.05$, fixing
\MethodABalancingAbbr{} and \MethodBAbbr{}.
Bottom and Top select the least and most PI-sensitive interaction steps.
(b) Cycle~3 retention, fixing \MethodASelectionAbbr{} and
\MethodABalancingAbbr{}.
Entries report success without PI (\%) as mean (SD) over three
decoding seeds; bold marks the highest mean in each column.}
    \label{tab:selection_retention_ablation}
    \vspace{5pt}
    \footnotesize
    \setlength{\tabcolsep}{0pt}
    \renewcommand{\arraystretch}{1.18}
    \newcommand{\resultcell}[2]{\begin{tabular}{@{}D{.}{.}{3.1}@{\hspace{1pt}}l@{}}#1 & {\scriptsize(#2)}\end{tabular}}
    \newcommand{\bestselectionresult}[2]{\begingroup\bfseries\boldmath\resultcell{#1}{#2}\endgroup}
    \begin{minipage}[t]{0.485\textwidth}
        \centering
        \textbf{(a) Selection strategy} \(r=1\)\par\smallskip
    \begin{tabular}{@{}>{\hspace{3pt}}p{0.34\linewidth}*{2}{>{\centering\arraybackslash}p{0.33\linewidth}}@{}}
        \toprule
        Selection & \textbf{ID} & \textbf{OOD} \\
        \midrule
        Random & \resultcell{49.7}{3.9} & \resultcell{45.8}{4.4} \\
        Bottom & \resultcell{40.1}{3.2} & \resultcell{40.4}{4.3} \\
        \rowcolor{black!5}
        \rule[-5pt]{0pt}{15pt}Top (ours) & \bestselectionresult{62.2}{2.0} & \bestselectionresult{57.3}{4.0} \\
        \bottomrule
    \end{tabular}
    \end{minipage}%
    \hfill
    \begin{minipage}[t]{0.485\textwidth}
        \centering
        \textbf{(b) Retention view} \(r=3\)\par\smallskip
    \begin{tabular}{@{}>{\hspace{3pt}}p{0.34\linewidth}*{2}{>{\centering\arraybackslash}p{0.33\linewidth}}@{}}
        \toprule
        Retention view & \textbf{ID} & \textbf{OOD} \\
        \midrule
        None & \resultcell{56.5}{3.0} & \resultcell{45.6}{1.6} \\
        Ordinary & \resultcell{72.9}{2.4} & \resultcell{65.9}{3.5} \\
        \rowcolor{black!5}
        \rule[-5pt]{0pt}{15pt}Privileged (ours) & \bestselectionresult{74.2}{1.4} & \bestselectionresult{67.2}{2.3} \\
        \bottomrule
    \end{tabular}
    \end{minipage}
\end{table}

\paragraph{PI Sensitivity Identifies Useful Interaction Steps.}
To test whether the selection criterion matters beyond sparsity,
we compare Random, Bottom, and Top selection at the same ratio
$\rho=0.05$, with \MethodABalancingAbbr{} and \MethodBAbbr{} fixed.
Top selection, corresponding to \MethodASelectionAbbr{}, improves
first-cycle success over Random by 12.5 and 11.5 percentage points
on ID and OOD, respectively, while Bottom selection performs worse
than Random on both splits
(Table~\ref{tab:selection_retention_ablation}(a)).
Thus, the benefit depends on prioritizing PI-sensitive steps,
rather than selecting an arbitrary subset of the same size.

\paragraph{Retention Improves Repeated-Deployment Performance.}
We compare no retention, ordinary-view retention, and privileged-view
retention while fixing \MethodASelectionAbbr{} and
\MethodABalancingAbbr{}.
The two retention objectives use the same set of interaction steps,
averaging scheme, and coefficient.
Both improve third-cycle deployment success substantially over
no retention, with privileged-view retention achieving the highest
results: 74.2\% on ID and 67.2\% on OOD
(Table~\ref{tab:selection_retention_ablation}(b)).
We next examine whether retaining the privileged view also better
preserves the PI-conditioned competence needed for the updated student to supervise learning in subsequent self-distillation cycles.

\begin{figure}[t]
    \centering
    \includegraphics[width=\linewidth]{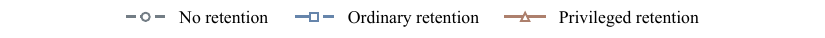}\par\nointerlineskip
    \begingroup
    \setkeys{Gin}{trim={0 3.5bp 0 0},clip}
    \alignedsubfloat{Ordinary view}{0.48363636\linewidth}{0.22556391}{0.04511278}{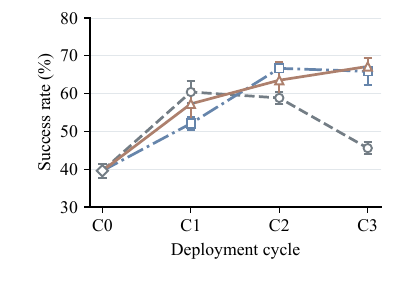}\hfill
    \alignedsubfloat{Privileged view}{0.48363636\linewidth}{0.22556391}{0.04511278}{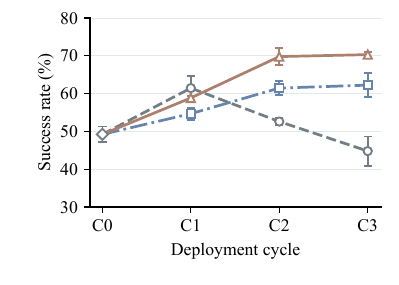}
    \endgroup
    \caption{\textbf{Privileged retention sustains PI-conditioned
task competence.}
OEL+\MethodAAbbr{} with Qwen3-4B on ALFWorld OOD tasks,
evaluated (a) without PI and (b) with PI.
Task initial states and the PI in (b) are fixed across checkpoints
and retention variants.
C0 denotes the shared initial model.
Points show means; error bars show $\pm1$ SD over three decoding seeds.}
    \label{fig:dual_view_evaluation}
\end{figure}

\begin{figure}[t]
    \centering
    \includegraphics[width=\linewidth]{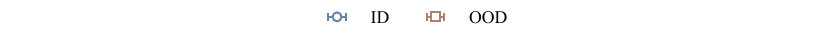}\par\nointerlineskip
    \alignedsubfloat{Selection ratio (Cycle 1)}{0.48363636\linewidth}{0.22556391}{0.04511278}{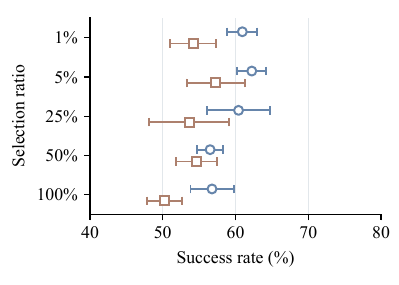}\hfill
    \alignedsubfloat{Retention weight (Cycle 3)}{0.48363636\linewidth}{0.22556391}{0.04511278}{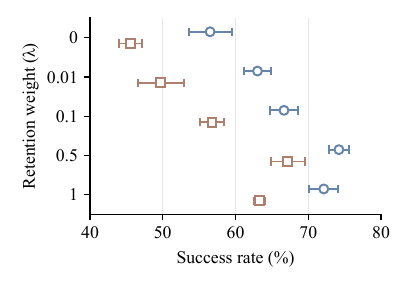}
    \caption{\textbf{Selection ratio and retention weight.}
OEL+\OverallMethod{} with Qwen3-4B on ALFWorld.
(a) Cycle~1 success at each selection ratio \(\rho\), fixing \(\lambda=0.5\).
(b) Cycle~3 success at each retention weight \(\lambda\), fixing \(\rho=0.05\);
each weight remains constant across cycles.
\MethodABalancingAbbr{} is enabled in both sweeps.
Points show means; horizontal error bars show \(\pm1\) SD over three decoding seeds.
}
    \label{fig:hyperparameters}
\end{figure}

\paragraph{Privileged Retention Preserves PI-Conditioned Competence.}
Without retention, PI-conditioned success falls from 61.5\% at
Cycle~1 to 44.8\% at Cycle~3.
With \MethodBAbbr{}, it instead increases from 58.9\% to 70.3\%
(Figure~\ref{fig:dual_view_evaluation}).
Ordinary-view retention reaches 62.2\% at Cycle~3, leaving an
8.1-percentage-point advantage for \MethodBAbbr{} under PI,
compared with a 1.3-point advantage without PI.
The larger separation in PI-conditioned performance supports
retaining the view in which the model provides supervision for
subsequent cycles.

\paragraph{Selection and Retention Hyperparameters.}
We evaluate the selection ratio after one cycle to measure
within-cycle learning, and the retention weight after three cycles
to measure its effect under repeated deployment.
With \MethodABalancingAbbr{} and \MethodBAbbr{} enabled and \(\lambda=0.5\),
selecting the top 1\%, 5\%, or 25\% of steps outperforms using all
steps in \(\mathcal U\) (\(\rho=1\)) on both splits (Figure~\ref{fig:hyperparameters}(a)).
The default \(\rho=0.05\) achieves the strongest first-cycle results,
improving over the \(\rho=1\) configuration by 5.4 and 7.0 percentage
points on ID and OOD.
With \(\rho=0.05\), third-cycle success increases as \(\lambda\) rises
from 0 to 0.5, while \(\lambda=1\) yields lower performance (Figure~\ref{fig:hyperparameters}(b)).
These sweeps favor selective distillation with a moderate retention weight, supporting the default settings used in our experiments.

\begin{figure}[!t]
\centering
\alignedsubfloat{Training cost}{0.48363636\linewidth}{0.22556391}{0.04511278}{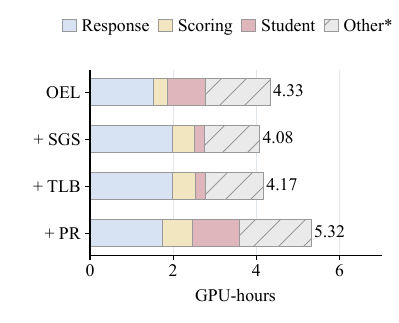}\hfill
\alignedsubfloat{AITZ accuracy}{0.48363636\linewidth}{0.22556391}{0.04511278}{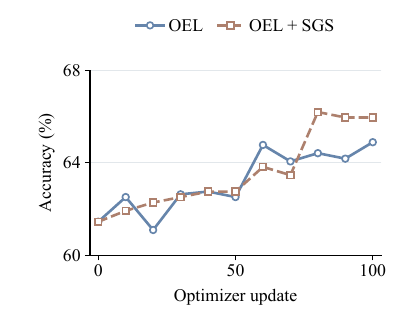}
\caption{\textbf{Training cost and AITZ accuracy.}
(a) Approximate Cycle~1 training-loop costs on ALFWorld using
eight H800 GPUs.
Components are added cumulatively to OEL; the final bar is
OEL+\OverallMethod{}.
Other* denotes residual training-loop cost.
(b) AITZ step-level action accuracy on a fixed subset.
OEL+\MethodASelectionAbbr{} retains a fixed top-80\% training subset.}
\label{fig:efficiency}
\label{fig:aitz_training_curves}
\end{figure}

\subsection{Efficiency and Generalization}

\paragraph{Training Cost--Performance Tradeoff.}
For Qwen3-4B on ALFWorld, \MethodASelectionAbbr{} improves the Cycle~1
training-loop cost--performance tradeoff (Figure~\ref{fig:efficiency}(a)).
It reduces student optimization cost by approximately 74\% and
total training-loop cost from 4.33 to 4.08 GPU-hours, while improving
first-cycle task success.
The total training-loop reduction is approximately 6\%. Response generation and sensitivity scoring account for most of the remaining cost. Adding \MethodABalancingAbbr{} increases total cost only slightly, from 4.08 to 4.17 GPU-hours, while student optimization remains at 0.24 GPU-hours.
The full \OverallMethod{} configuration additionally retains
privileged-view predictions at all interaction steps in \(\mathcal U\), bringing the
measured cost to 5.32 GPU-hours, or 1.23 times that of OEL.
Thus, selection offers a lower-cost improvement, while the complete
method preserves PI-conditioned competence to sustain task performance over multiple deployment cycles.

\paragraph{PI Sensitivity as an Offline Multimodal Data Filter.}
We further apply \MethodASelectionAbbr{} to an existing dataset of
multimodal agent trajectories on AITZ.
Using PI sensitivity computed by the initial model, we retain the
global top 80\% of interaction steps and keep this subset fixed
throughout training.
Both methods share the same initialization, 100-update schedule, and OEL objective.

OEL+\MethodASelectionAbbr{} achieves higher action accuracy at
the final three evaluations and finishes at 65.95\%, compared with
64.89\% for OEL.
This result extends PI-sensitive selection beyond iterative
text-agent training to static offline filtering for a multimodal
agent, improving final action accuracy by 1.06 percentage points.
These gains are obtained with the same update budget and a subset fixed before training, supporting PI sensitivity as a criterion for filtering offline agent trajectories.

%% file: sections/05_conclusion.tex
\section{Conclusion}

We introduced \OverallMethod{}, a plug-in augmentation for iterative experience-based self-distillation.
It combines PI-sensitive selection and trajectory-balanced distillation with privileged retention to improve learning within each cycle and preserve PI-conditioned behavior across cycles.
Experiments on ALFWorld and TextCraft show that \OverallMethod{} improves SDPO and OEL across model scales and sustains gains over three deployment cycles.
Ablations and evaluations with fixed privileged information support the complementary roles of selection and retention.
Sensitivity-guided offline filtering also improves action prediction for multimodal agents on AITZ.
These findings highlight the importance of improving the deployed policy while preserving the PI-conditioned competence it will use as the teacher in subsequent self-distillation cycles.

%% file: sections/06_appendix.tex
\clearpage
\appendix
\appendixtableofcontents
\clearpage
% APP-GAP R01: user-owned bibliography and verified citation keys.
% Completion tracker: writing/2026-09-06_appendix_completion_checklist.md
\input{sections/appendix/a_experimental_setup}
\clearpage
\input{sections/appendix/b_implementation}
\clearpage
\input{sections/appendix/c_robustness}
\clearpage
\input{sections/appendix/d_analysis}
\clearpage
\input{sections/appendix/g_prompts}

%% file: sections/appendix/a_experimental_setup.tex
\section{Experimental Setup}
\label{app:experimental_settings}
\subsection{Benchmarks and Data Splits}
\label{app:benchmarks}
Text experiments use Qwen3-4B and Qwen3-8B with thinking disabled;
AITZ uses Qwen3-VL-4B-Instruct. Table~\ref{tab:benchmark_protocol}
summarizes the test sets, held fixed across methods and cycles.

\begin{table}[htbp]
\centering\small
\caption{Evaluation scope per decoding seed.}
\label{tab:benchmark_protocol}
\renewcommand{\arraystretch}{1.15}
\begin{tabular}{@{}>{\raggedright\arraybackslash}p{.20\linewidth}
>{\raggedright\arraybackslash}p{.56\linewidth}
>{\raggedright\arraybackslash}p{.14\linewidth}@{}}
\toprule
Benchmark & Test set & Metrics \\
\midrule
ALFWorld & 128 ID and 128 OOD episodes & Success rate \\
TextCraft & Full official test set: 100 episodes & Success rate \\
AITZ & Fixed monitoring subset: 101 episodes, 843 non-\texttt{stop} steps & Accuracy \\
\bottomrule
\end{tabular}
\end{table}

\subsection{Training Data and Cycles}
\label{app:collection}
Text agents use three cycles with the budgets in Table~\ref{tab:training_configuration},
carrying forward final model parameters and resetting optimizer and random states.
Each cycle samples task IDs without replacement and collects one trajectory
per task; recurring tasks receive fresh trajectories. Methods of the same model size share the
Cycle~1 corpus and summaries; later cycles use each method's own preceding
final checkpoint to collect a fresh corpus, replacing the previous one.

AITZ uses 1,728 externally collected, complete CoAT training trajectories
from Aguvis Stage-2~\citep{pmlr-v267-xu25ae}, each with at most 20 steps
and no unsupported actions. We use their screenshots
and logged actions, without the original GPT-generated semantic annotations.
A fixed seed samples and orders 1,600 trajectories without replacement,
shared by all training runs. The initial policy prepares diagnostic attempts
and trajectory summaries before optimization. OEL and OEL + \MethodASelectionAbbr{}
follow the same single-cycle, 100-update schedule, with 16 source trajectories per update.
The filtered run retains a fixed global top 80\% of steps before training.
Only retained steps contribute to its loss; their original
histories and summaries remain available.

\subsection{Evaluation and Reporting}
\label{app:evaluation}
\label{app:gui_protocol}
Text tasks evaluate each cycle's final checkpoint, without intermediate
validation or checkpoint selection; ReAct is the unchanged initial policy.
Success rate uses a 30-step interaction limit. Invalid actions consume a
step, and the agent continues from environment feedback without regenerating
the response. ALFWorld tasks and initial states are fixed across decoding seeds.
Final results for the text benchmarks report mean and sample standard deviation
across three decoding seeds, measuring decoding variability.

For AITZ, we monitor both methods on a fixed random subset of 101 test episodes
every ten updates using one
fixed decoding seed. Monitoring does not guide hyperparameter changes,
early stopping, or checkpoint selection.

Within the fixed monitoring subset, AITZ evaluation excludes steps
whose reference action is \texttt{stop}.
At each remaining logged step, the agent receives the
task instruction, current screenshot, and ground-truth prior actions.
Predictions are scored against reference actions without execution or
insertion into later histories. Exact-match (EM) accuracy follows the official screen-wise rule
for the category and its associated arguments: target, scroll direction,
text, or control command, as applicable. Matching follows the scorer's
criteria rather than literal equality of raw arguments.
Accuracy averages over all non-\texttt{stop} evaluation steps;
invalid predictions score zero. Complete inputs are in Appendix~\ref{app:prompts}.

Evaluation uses temperature \(0.4\), top-\(p=1.0\), and no top-\(k\)
truncation (\(k=-1\)). Generation is capped at 1,024 tokens for ALFWorld,
TextCraft, and AITZ. Generation settings and the seeds for each
evaluation protocol are matched across methods and checkpoints.
% APP-GAP A03: verify evaluator versions and final evaluation manifests.
% APP-GAP E01: verify the actual AITZ scorer version against the run configuration.

%% file: sections/appendix/b_implementation.tex
\section{Implementation and Baseline Details}
\label{app:implementation}
This section specifies experience construction, the optimization procedure,
and baseline adaptations under the protocol in Appendix~\ref{app:experimental_settings}.

\paragraph{Training infrastructure.}
Our training framework is built on slime,\footnote{\url{https://github.com/THUDM/slime}}
with Megatron-LM\footnote{\url{https://github.com/NVIDIA/Megatron-LM}} for training
and SGLang\footnote{\url{https://github.com/sgl-project/sglang}} for generation.
We use Python 3.12.3 and PyTorch 2.11.0 with CUDA 12.9.
Model parameters use BF16, while optimizer states and main gradients use FP32.
Megatron uses data parallelism of degree 8 and tensor, pipeline, and context
parallelism of degree 1. Each SGLang engine uses tensor parallelism of degree 1.

\subsection{Inputs and Experience Construction}
\label{app:inputs}
The ordinary input contains the task and interaction history. PI adds a
summary \(c_\tau\) extracted from the associated complete trajectory.
Text-task extraction uses a task snapshot, executed actions, resulting
observations, and the trajectory outcome; AITZ additionally uses the cycle-initial
student's diagnostic attempts. The trajectory itself is not appended
to the teacher input.

Each PI-based method uses its frozen starting policy to extract one accepted
summary per trajectory, paired with its source histories and cached throughout
training. Text-task summaries refresh each cycle; AITZ prepares them once.
Extraction rules and templates are shared, although different policies may
produce different summaries. Appendices~\ref{app:collection}
and~\ref{app:prompts} specify data sharing and complete templates.
Deployment evaluation uses ordinary input.

\subsection{Selection, Retention, and Optimization}
\label{app:optimization}
Algorithm~\ref{alg:method_training} summarizes the full text-task integration using the objectives
in Section~\ref{sec:joint_optimization}. Selection operates over the full
update batch, with a fixed tie order and no trajectory-coverage completion.
Scoring and both losses use the same response prefixes and valid-token masks;
responses, scores, and selected indices are fixed during differentiation.
We compute teacher JSD and both distillation and retention KL using a
Top-20-plus-residual-tail approximation. At each response prefix, the current
ordinary-view student selects the top 20 token IDs; all compared
distributions use these same IDs and pool their remaining probability mass
into one tail category.

Distillation covers selected interaction steps; retention covers all steps in \(\mathcal U\).
Each loss averages within trajectories and then over the full sampled batch,
with empty supports contributing zero. We neither renormalize over surviving
trajectories nor adjust \(\lambda\) with coverage. Thus, \(\rho=1\) need not
recover the baseline's uniform average over steps.

\begin{algorithm}[htbp]
\caption{One internalization cycle of \OverallMethod{}}
\label{alg:method_training}
\small
\algrenewcommand\algorithmicrequire{\textbf{Input:}}
\algrenewcommand\algorithmicensure{\textbf{Output:}}
\begin{algorithmic}[1]
\Require Trajectories \(\mathcal D_r\), starting policy \(\pi_r\), selection ratio \(\rho\), retention weight \(\lambda\), updates \(M\)
\Ensure Final student \(\pi_{r+1}\), evaluated with ordinary input.
\State Initialize student \(\pi^{\mathrm S}\) and teacher \(\pi^{\mathrm T}\) from \(\pi_r\); freeze teacher
\State Prepare per-trajectory summaries \(c_\tau\) using the base-method interface
\For{\(u=1,\ldots,M\)}
    \State Take the next trajectory batch \(\mathcal B\) from the cycle's data schedule
    \State Obtain base-method responses \(z_{\tau,t}\); form \(\mathcal U\)
    \State Initialize \(\mathcal C_\rho\leftarrow\emptyset\), \(\mathcal L_{\mathrm{sel}}\leftarrow0\), \(\mathcal L_{\mathrm{ret}}\leftarrow0\)
    \If{\(\mathcal U\neq\emptyset\)}
        \State Score \(\mathcal U\) with frozen-teacher JSD (Equation~\ref{eq:privileged-sensitivity})
        \State \(\mathcal C_\rho\leftarrow\operatorname{TopK}(\mathcal U;s,\lceil\rho|\mathcal U|\rceil)\), using fixed tie order
        \State Compute \(\mathcal L_{\mathrm{sel}}\) via Equation~\ref{eq:selected-distillation-objective}, applying base-method token corrections before reduction
        \State Compute \(\mathcal L_{\mathrm{ret}}\) via Equation~\ref{eq:retention-objective}
    \EndIf
    \State Optimize only student with \(\mathcal L_{\mathrm{sel}}+\lambda\mathcal L_{\mathrm{ret}}\)
\EndFor
\State \Return Final student \(\pi_{r+1}\), evaluated with ordinary input
\end{algorithmic}
\end{algorithm}

All trained methods use AdamW with constant learning rate \(10^{-6}\)
and no warmup. Table~\ref{tab:training_configuration} gives the
trajectory budgets and \OverallMethod{} hyperparameters.

\begin{table}[htbp]
\centering\small
\caption{Training settings per cycle (one cycle for AITZ, three for text tasks).
Batch size counts source trajectories before step filtering;
\(\rho\) and \(\lambda\) denote selection ratio and retention weight.}
\label{tab:training_configuration}
\begin{tabular}{@{}lrrrrr@{}}
\toprule
Benchmark & Updates & Batch size & Trajectories & \(\rho\) & \(\lambda\) \\
\midrule
ALFWorld & 30 & 32 & 960 & 0.05 & 0.5 \\
TextCraft & 30 & 8 & 240 & 0.25 & 1.0 \\
AITZ (OEL) & 100 & 16 & 1,600 & -- & -- \\
AITZ (OEL + SGS) & 100 & 16 & 1,600 & 0.80 & -- \\
\bottomrule
\end{tabular}
\end{table}

\paragraph{Offline filtering on AITZ.}
We score initial-policy ordinary responses using
Equation~\ref{eq:privileged-sensitivity} and retain the fixed global top 80\%
of steps for OEL + \MethodASelectionAbbr{}. The filtered run preserves the
OEL loss, without \MethodABalancingAbbr{} or \MethodBAbbr{}.
OEL~\citep{ye2026online} distills the frozen privileged teacher along fresh
ordinary-student responses at each update, without changing this selection.

\paragraph{Ordinary-view retention control.}
The Ordinary control in Table~\ref{tab:selection_retention_ablation}(b)
matches the ordinary student to the ordinary frozen cycle-start teacher.
KL direction, set of interaction steps, token and trajectory averaging, and coefficient
match privileged retention; only the conditioning views change.

\subsection{Baseline Adaptations}
\label{app:baselines}
Table~\ref{tab:baseline_contract} summarizes the baselines.
AITZ uses the OEL filtering comparison above.
ReAct~\citep{yao2023reactsynergizingreasoningacting} is the unchanged initial policy.
All training uses fixed logged histories without executing new environment actions.

\begin{table}[htbp]
\centering\small
\caption{Baseline response sources and training objectives.
All methods use ordinary input at evaluation. SDPO in the result tables
refers to the offline adaptation below.}
\label{tab:baseline_contract}
\begin{tabular}{@{}>{\raggedright\arraybackslash}p{.17\linewidth}
>{\raggedright\arraybackslash}p{.43\linewidth}
>{\raggedright\arraybackslash}p{.30\linewidth}@{}}
\toprule
Method & Response source & Training objective \\
\midrule
ReAct & Initial policy & No update \\
RFT & Successful logged responses & Cross-entropy \\
GRPO & Policy-generated candidate groups & Clipped objective + KL \\
EPD & Cached privileged-teacher responses & Cross-entropy \\
Offline SDPO & Cached cycle-start student responses & Distillation with correction \\
OEL & Fresh current-student responses & Distillation \\
\bottomrule
\end{tabular}
\end{table}

\paragraph{RFT and success filtering.}
RFT~\citep{yuan2023scalingrelationshiplearningmathematical} fits successful
logged responses by cross-entropy under ordinary histories. RFT and GRPO
reuse successes from the current cycle's corpus as needed to match the
update budget, without additional environment attempts.

\newpage
\paragraph{GRPO: offline action matching.}
GRPO~\citep{shao2024deepseekmathpushinglimitsmathematical} samples eight responses
per successful logged step. Reward is one if the normalized action type and
all arguments match the log, and zero otherwise, including invalid actions.
Reasoning text is ignored. Group advantages are
\[
 A_i=\frac{r_i-\bar r}{s_r+10^{-6}},
 \qquad s_r^2=\frac{1}{7}\sum_{j=1}^{8}(r_j-\bar r)^2.
\]
No additional batch/token advantage whitening is applied. PPO ratios are
clipped to \([0.8,1.28]\). The frozen base Qwen3-4B or Qwen3-8B reference,
matching the trained model's size, supplies a \texttt{low\_var\_kl} auxiliary
loss with \(\beta=0.01\), excluded from rewards and advantages. Entropy has zero weight.

\paragraph{EPD.}
EPD~\citep{gou2026sample} caches the frozen cycle-start teacher's complete,
summary-conditioned responses before training, then fits these targets
by ordinary-view cross-entropy.

\paragraph{Offline SDPO.}
\label{app:offline_sdpo}
Offline SDPO~\citep{hubotter2026reinforcement} uses newly generated cycle-start
ordinary responses, one per logged history, as cached KL prefixes throughout
the cycle. Its baseline averages valid-token KL within steps, then uniformly
across steps in \(\mathcal U\).

Following SDPO~\citep[Appendix~A.4]{hubotter2026reinforcement}, token-level
truncated importance sampling (TIS) corrects distillation. Both current-student/old-actor
and old-actor/cycle-start-policy ratios are capped at 2. The old actor refreshes
before each update; the teacher stays frozen for the cycle. Baseline and
augmented SDPO share this correction.

%% file: sections/appendix/c_robustness.tex
\section{Detailed Experimental Results}
\label{app:robustness}
\label{app:multimodal_cost}
The main results and primary ablations are reported in
Tables~\ref{tab:main_results}--\ref{tab:selection_retention_ablation}.
This appendix complements the main experiments with full hyperparameter
results, AITZ training dynamics, and a breakdown of training-loop costs.

\subsection{Hyperparameter Ablations}
\label{app:hyperparameters}

Table~\ref{tab:hyperparameter_results} provides the numerical results for
the two sweeps in Figure~\ref{fig:hyperparameters}.
Both use OEL+\OverallMethod{} with Qwen3-4B on ALFWorld, changing one
parameter at a time from the default \(\rho=0.05,\lambda=0.5\).
Training budgets and optimization settings follow
Table~\ref{tab:training_configuration}; evaluation follows
Appendix~\ref{app:evaluation}.

\begin{table}[htbp]
\centering\footnotesize
\caption{\textbf{Complete hyperparameter ablation results.}
Success rates (\%) are reported as mean (sample standard deviation)
over three decoding seeds, with 128 episodes per split in each evaluation.
Shading marks the default parameter in each sweep.
The two panels report different deployment cycles.}
\label{tab:hyperparameter_results}
\begingroup
\setlength{\tabcolsep}{0pt}
\renewcommand{\arraystretch}{1.18}
\newcommand{\hyperresult}[2]{\begin{tabular}{@{}D{.}{.}{3.1}@{\hspace{1pt}}l@{}}#1 & {\scriptsize(#2)}\end{tabular}}
\begin{minipage}[t]{0.485\textwidth}
\centering
\textbf{(a) Selection ratio} \(r=1\)\par\smallskip
\begin{tabular}[t]{@{}>{\hspace{3pt}}p{0.34\linewidth}*{2}{>{\centering\arraybackslash}p{0.33\linewidth}}@{}}
\toprule
Top-\(\rho\) & \textbf{ID} & \textbf{OOD} \\
\midrule
1\% & \hyperresult{60.9}{2.1} & \hyperresult{54.2}{3.2} \\
\rowcolor{black!5}
5\% & \hyperresult{62.2}{2.0} & \hyperresult{57.3}{4.0} \\
25\% & \hyperresult{60.4}{4.3} & \hyperresult{53.6}{5.5} \\
50\% & \hyperresult{56.5}{1.8} & \hyperresult{54.7}{2.8} \\
100\% & \hyperresult{56.8}{3.0} & \hyperresult{50.3}{2.4} \\
\bottomrule
\end{tabular}
\end{minipage}%
\hfill
\begin{minipage}[t]{0.485\textwidth}
\centering
\textbf{(b) Retention weight} \(r=3\)\par\smallskip
\begin{tabular}[t]{@{}>{\hspace{3pt}}p{0.34\linewidth}*{2}{>{\centering\arraybackslash}p{0.33\linewidth}}@{}}
\toprule
\(\lambda\) & \textbf{ID} & \textbf{OOD} \\
\midrule
0 & \hyperresult{56.5}{3.0} & \hyperresult{45.6}{1.6} \\
0.01 & \hyperresult{63.0}{1.8} & \hyperresult{49.7}{3.2} \\
0.1 & \hyperresult{66.7}{2.0} & \hyperresult{56.8}{1.6} \\
\rowcolor{black!5}
0.5 & \hyperresult{74.2}{1.4} & \hyperresult{67.2}{2.3} \\
1 & \hyperresult{72.1}{2.0} & \hyperresult{63.3}{0.8} \\
\bottomrule
\end{tabular}
\end{minipage}
\endgroup
\end{table}

\paragraph{Selection ratio.}
With \(\lambda=0.5\), the sweep changes only the fraction of interaction steps in \(\mathcal U\) used
for distillation. \MethodABalancingAbbr{} remains enabled, and
\MethodBAbbr{} covers all steps in \(\mathcal U\) at the fixed weight.
Consequently, Top-100\% retains both trajectory balancing and retention;
it is not the unaugmented OEL baseline.

\paragraph{Retention weight.}
With \(\rho=0.05\), each setting starts from the same base model and keeps its weight fixed
across three cycles, carrying forward its own final student between cycles.
The table reports the third-cycle final checkpoints.
At \(\lambda=0\), retention is removed while selection and
\MethodABalancingAbbr{} remain enabled.
The error bars in the main figure and the standard deviations here
describe decoding variability, rather than variation across independent
training runs.

\clearpage
\subsection{AITZ Training Dynamics}
\label{app:multimodal_results}
Figure~\ref{fig:aitz_training_curves}(b) reports accuracy every ten updates
on 843 non-\texttt{stop} steps from a fixed random subset of 101 test episodes,
using one decoding seed. At update 100, OEL reaches 64.89\% accuracy and
OEL + \MethodASelectionAbbr{} (Top-80\% steps) reaches 65.95\%.
The monitoring subset is not used for model selection.
Appendix~\ref{app:evaluation} specifies the shared scoring rules.

\FloatBarrier
\subsection{Computational Cost}
\label{app:cost_accounting}
Table~\ref{tab:runtime_breakdown} provides the numerical breakdown for
Figure~\ref{fig:efficiency}(a): approximate Cycle~1 training-loop costs on
ALFWorld with the default OEL integration and eight NVIDIA H800 GPUs.
GPU-hours are elapsed hours multiplied by eight, counting all
allocated GPUs without utilization weighting. Auxiliary JSD diagnostics
are excluded; required \MethodASelectionAbbr{} scoring and privileged-view
forward and backward passes for \MethodBAbbr{} over all steps in \(\mathcal U\) are included.
Other is the residual between total loop cost and the three compute
components, including model switching, synchronization, and scheduling
waits. It is not separately profiled; the measured scope is the training
loop, not the full deployment cycle.

\begin{table}[htbp]
\centering\small
\caption{Training-loop cost on ALFWorld (Cycle~1, GPU-hours on eight H800 GPUs).
Scoring includes selection; Other is computed as the residual.
Values are approximate; totals are computed before rounding.}
\label{tab:runtime_breakdown}
\setlength{\tabcolsep}{0pt}
\renewcommand{\arraystretch}{1.18}
\begin{tabular}{@{}>{\raggedright\arraybackslash}p{0.30\linewidth}*{5}{>{\centering\arraybackslash}p{0.14\linewidth}}@{}}
\toprule
\textbf{Configuration} & \textbf{Response} & \textbf{Scoring} & \textbf{Optimization} & \textbf{Other} & \textbf{Total} \\
\midrule
OEL & 1.53 & 0.33 & 0.91 & 1.56 & 4.33 \\
OEL + \MethodASelectionAbbr{} & 1.97 & 0.55 & 0.24 & 1.32 & 4.08 \\
OEL + \MethodASelectionAbbr{} + \MethodABalancingAbbr{} & 1.99 & 0.56 & 0.24 & 1.39 & 4.17 \\
\OverallMethod{} (+ \MethodBAbbr{}) & 1.73 & 0.72 & 1.13 & 1.73 & 5.32 \\
\bottomrule
\end{tabular}
\end{table}

%% file: sections/appendix/d_analysis.tex
\section{Analysis and Discussion}
\label{app:additional_diagnostics}
This appendix discusses privileged retention and provides the protocols
for dual-view evaluation and PI-sensitivity diagnostics.
All empirical analyses use ALFWorld and Qwen3-4B.
Here, PI consists of trajectory-derived experience summaries.

\subsection{Privileged Retention versus Reference KL Regularization}
\label{app:retention_reference_kl}

\paragraph{Regularization target.}
\MethodBAbbr{} applies KL regularization to preserve behavior conditioned
on privileged information (PI).
Reference KL in GRPO penalizes deviation from a reference policy during
reward optimization~\citep{shao2024deepseekmathpushinglimitsmathematical}.
Our objective targets the model's ability to use PI: the updated student
becomes the next cycle's teacher and supplies privileged-view supervision.

\paragraph{Conditioning view.}
Our ordinary-view control and \MethodBAbbr{} match the KL direction,
set of interaction steps, loss averaging, and coefficient (\(\lambda=0.5\)).
Each configuration freezes its own cycle-start policy as the teacher.
Omitting response prefixes, the two penalties compare
\[
\begin{aligned}
\text{Ordinary:}\quad &
D_{\mathrm{KL}}\!\left(
\pi^{\mathrm S}(\cdot\mid h)
\,\middle\|\,\pi^{\mathrm T}(\cdot\mid h)\right),\\
\text{Privileged:}\quad &
D_{\mathrm{KL}}\!\left(
\pi^{\mathrm S}(\cdot\mid h,c)
\,\middle\|\,\pi^{\mathrm T}(\cdot\mid h,c)\right),
\end{aligned}
\]
where \(h\) is the interaction history and \(c\) is PI.
Although the views share parameters, constraining predictions given
\(h\) does not directly constrain predictions given \((h,c)\).
\MethodBAbbr{} applies the latter constraint at all steps in \(\mathcal U\)
(Equation~\ref{eq:privileged-retention}).

\paragraph{Functional evidence.}
Figure~\ref{fig:dual_view_evaluation} separates ordinary-view performance
from the ability to benefit from Experience.
For OEL+\MethodAAbbr{} at Cycle~3 on OOD tasks, ordinary-view retention
improves both input-view means over no retention, but its success decreases
when Experience is supplied.
Under \MethodBAbbr{}, Experience instead raises mean success by
3.1 percentage points.
This contrast supports explicitly retaining Experience-conditioned behavior.

\paragraph{Retention strength.}
The \MethodBAbbr{} weight sweep in Figure~\ref{fig:hyperparameters}(b)
fixes \(\rho=0.05\) throughout three cycles.
Increasing \(\lambda\) from 0.01 to 0.5 raises Cycle~3 mean success
from 63.0\% to 74.2\% on ID and from 49.7\% to 67.2\% on OOD;
increasing it to 1 lowers both means
(Appendix~\ref{app:hyperparameters}).
Our GRPO baseline also uses a KL coefficient of 0.01, but coefficient
values depend on the primary objective and loss normalization;
their numerical sizes alone do not determine the strength of regularization.

\subsection{Dual-View Evaluation Details}
\label{app:dual_view_evaluation}

\paragraph{Evaluation protocol.}
We evaluate ten checkpoints: the shared initial model and the final models
from Cycles~1--3 of OEL+\MethodAAbbr{} with no retention, ordinary-view
retention, or \MethodBAbbr{}.
Each is evaluated with and without Experience on the same 128 OOD tasks,
using matched initial states and three decoding seeds.
Thinking is disabled; decoding uses temperature \(0.4\), top-\(p=1\),
no top-\(k\) truncation, and at most 1,024 tokens per response,
with a 30-step interaction limit
(Appendix~\ref{app:evaluation}).
Table~\ref{tab:dual_view_results} reports the complete results.

\paragraph{Fixed Experience.}
To hold guidance constant, the frozen initial model collects one reference
episode per task in the ordinary view and summarizes it using the
success/failure prompts in Appendix~\ref{app:prompts}.
Both successful and unsuccessful episodes are retained.
The resulting 128 task-specific Experiences remain fixed across
configurations, checkpoints, and decoding seeds, rather than being
refreshed at each training cycle.
Each evaluation episode restarts from its task-specific initial state
and builds its own action history; only the privileged view
receives the fixed summary at every interaction step.
This protocol compares checkpoint competence under fixed guidance;
it does not assess summary generation or subsequent teaching effects.

\begin{table}[htbp]
\centering\small
\caption{\textbf{Task success under both input views.}
Mean success rates (\%) and sample standard deviations over three decoding
seeds on 128 ALFWorld OOD tasks. All trained configurations use
OEL+\MethodAAbbr{} and differ only in retention; the shared initial
checkpoint is listed once.}
\label{tab:dual_view_results}
\begingroup
\setlength{\tabcolsep}{0pt}
\renewcommand{\arraystretch}{1.18}
\begin{tabular}{@{}p{0.40\linewidth}
    >{\centering\arraybackslash}p{0.12\linewidth}
    >{\centering\arraybackslash}p{0.22\linewidth}
    >{\centering\arraybackslash}p{0.26\linewidth}@{}}
\toprule
Retention & Cycle & Ordinary view & Privileged view \\
\midrule
Shared initial model & 0 & 39.6 (1.8) & 49.2 (2.1) \\
\midrule
\multirow{3}{*}{None} & 1 & 60.4 (3.0) & 61.5 (3.3) \\
 & 2 & 58.9 (1.6) & 52.6 (0.9) \\
 & 3 & 45.6 (1.6) & 44.8 (3.9) \\
\addlinespace
\multirow{3}{*}{Ordinary} & 1 & 52.1 (1.6) & 54.7 (1.6) \\
 & 2 & 66.7 (0.9) & 61.5 (1.8) \\
 & 3 & 65.9 (3.5) & 62.2 (3.3) \\
\addlinespace
\multirow{3}{*}{Privileged (\MethodBAbbr{})} & 1 & 57.3 (4.0) & 58.9 (0.5) \\
 & 2 & 63.5 (4.7) & 69.8 (2.3) \\
 & 3 & 67.2 (2.3) & 70.3 (0.8) \\
\bottomrule
\end{tabular}
\endgroup
\end{table}

\FloatBarrier
\subsection{PI-Sensitivity Measurement and Dynamics}
\label{app:view_drift}
\label{app:diagnostic_protocol}

\begin{figure}[!t]
\centering
\alignedsubfloat{OEL component combinations}{\linewidth}{0.125}{0.015}{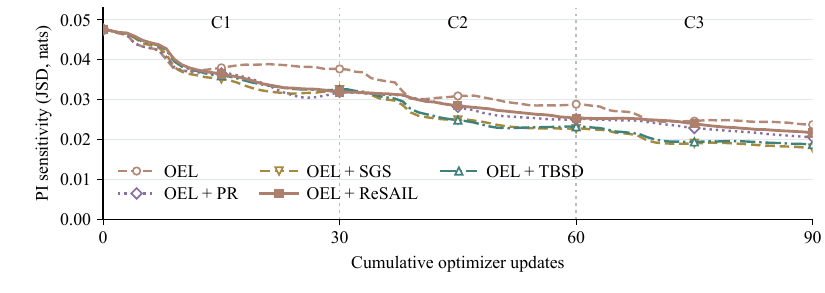}
\par\smallskip
\alignedsubfloat{SDPO integration and EPD}{\linewidth}{0.125}{0.015}{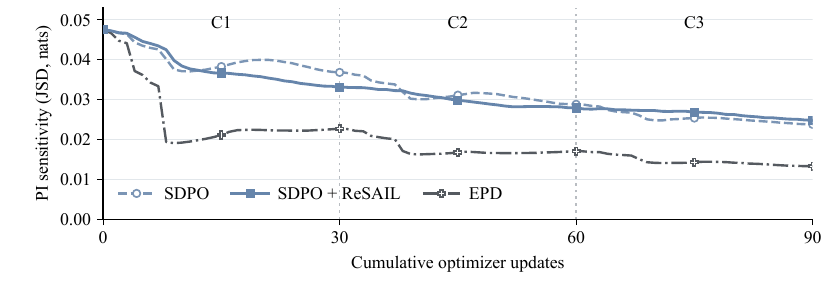}
\caption{\textbf{Full PI-sensitivity dynamics across three deployment cycles.}
Current-student JSD on the fixed ALFWorld/Qwen3-4B probe, using
response-token averaging. (a) OEL and component combinations.
(b) SDPO with and without \OverallMethod{}, plus EPD.
All observations are shown without smoothing across 90 updates;
vertical lines mark cycle boundaries and markers are spaced for readability.}
\label{fig:all_sensitivity_dynamics}
\end{figure}

Figure~\ref{fig:all_sensitivity_dynamics} shows the complete, unsmoothed
PI-sensitivity trajectories for all eight configurations.

\paragraph{Fixed probe.}
We use 16 logged trajectories comprising 315 interaction steps.
At each cycle's initialization and after every update, we evaluate both
views of the current student on the same logged responses.
At each response prefix, vocabulary support is fixed to the initial
model's ordinary-view Top-20 tokens plus one residual-tail category.
This fixed support keeps diagnostic comparisons consistent across checkpoints;
training uses the current-student support described in Appendix~\ref{app:optimization}.
PI sensitivity is the JSD between the ordinary- and privileged-view
distributions, reported in nats and averaged over all response tokens;
longer responses therefore receive more weight.
The probe is a fixed diagnostic sample, not an independent task-held-out
set, and all curves describe one training seed.

\FloatBarrier

Sensitivity generally decreases, with local reversals less pronounced
under \OverallMethod{} than under OEL and SDPO.
Adding \MethodBAbbr{} to OEL+\MethodAAbbr{} does not uniformly raise
endpoint JSD across cycles.
These curves describe distributional responses to Experience;
functional competence is evaluated separately in Section~\ref{sec:ablation_studies}
and Appendix~\ref{app:dual_view_evaluation}.

%% file: sections/appendix/g_prompts.tex
\clearpage
\section{Prompt Templates}
\label{app:prompts}
\begingroup\raggedbottom
\lstset{aboveskip=0pt,belowskip=0pt}
Each box assembles a complete model call, including messages, tools, and
image positions. Runtime fields such as \texttt{<CURRENT\_OBSERVATION>} are
placeholders; tags such as \texttt{<thinking>} and \texttt{<action>} are literal.
\texttt{<IMAGE: ...>} marks image positions, and square-bracketed role and
conditional labels are display annotations.
Ellipses denote repeated records or runtime fields, not omitted instructions.
Ordinary inputs omit the privileged insertion; text privileged inputs include
one outcome-matched summary. Native thinking is disabled in the chat template;
the explicitly requested response formats still apply.

\subsection{Text Agent Prompts}
All previous observations and executed actions are serialized into the
current user message, separated by spaces. Earlier reasoning is omitted.
The full episode history is retained, up to the 30-step text-task limit.
For ALFWorld, the task description is extracted from the initial
observation, and \texttt{help} is removed from the admissible actions.
For TextCraft, episode-level crafting commands are reinserted when available.

\begin{promptlisting}{ALFWorld: first-step input}
[USER MESSAGE]
You are an expert agent operating in the ALFRED Embodied Environment.
Your current observation is: <CURRENT_OBSERVATION>
Your admissible actions of the current situation are: [
 '<ACTION_1>'
 '<ACTION_2>'
 ...
 '<ACTION_N>'].

Now it's your turn to take an action.
You should first reason step-by-step about the current situation. This reasoning process MUST be enclosed within <thinking> </thinking> tags.
Once you've finished your reasoning, you should choose an admissible action for current step and present it within <action> </action> tags.

[IF PRIVILEGED VIEW AND SUCCESSFUL TRAJECTORY]
A successful trajectory for the current task:

Guidance summary:
- Minimal plan: <...>
- Critical actions: <...>
- Checks: <...>
- Avoid: <...>

Use this information as a reference and continue solving the original task.

[ELSE IF PRIVILEGED VIEW AND FAILED TRAJECTORY]
A failed trajectory for the current task:

Failure analysis:
- Failure diagnosis: <...>
- Useful evidence: <...>
- Corrected plan: <...>
- Avoid: <...>

Use this information as a reference and continue solving the original task.
[END CONDITIONAL INSERT]
\end{promptlisting}

\clearpage
\begin{promptlisting}{ALFWorld: later-step input}
[USER MESSAGE]
You are an expert agent operating in the ALFRED Embodied Environment. Your task is to: <TASK_DESCRIPTION>
Prior to this step, you have already taken <STEP_COUNT> step(s). Below are the most recent <HISTORY_LENGTH> observations and the corresponding actions you took: [Observation 1: '<OBSERVATION_1>', Action 1: '<ACTION_1>'] [Observation 2: '<OBSERVATION_2>', Action 2: '<ACTION_2>'] ... [Observation K: '<OBSERVATION_K>', Action K: '<ACTION_K>']
You are now at step <CURRENT_STEP> and your current observation is: <CURRENT_OBSERVATION>
Your admissible actions of the current situation are: [
 '<ACTION_1>'
 '<ACTION_2>'
 ...
 '<ACTION_N>'].

Now it's your turn to take an action.
You should first reason step-by-step about the current situation. This reasoning process MUST be enclosed within <thinking> </thinking> tags.
Once you've finished your reasoning, you should choose an admissible action for current step and present it within <action> </action> tags.

[IF PRIVILEGED VIEW AND SUCCESSFUL TRAJECTORY]
A successful trajectory for the current task:

Guidance summary:
- Minimal plan: <...>
- Critical actions: <...>
- Checks: <...>
- Avoid: <...>

Use this information as a reference and continue solving the original task.

[ELSE IF PRIVILEGED VIEW AND FAILED TRAJECTORY]
A failed trajectory for the current task:

Failure analysis:
- Failure diagnosis: <...>
- Useful evidence: <...>
- Corrected plan: <...>
- Avoid: <...>

Use this information as a reference and continue solving the original task.
[END CONDITIONAL INSERT]
\end{promptlisting}

\clearpage
\begin{promptlisting}{TextCraft: first-step input}
[USER MESSAGE]
You are an expert agent operating in TextCraft, a text-only Minecraft crafting environment.
Your current observation is:
<CURRENT_OBSERVATION>

Valid actions are:
- get N item
- inventory
- craft N target item using N ingredient, N ingredient

Use only the crafting commands listed in the observation. Always include quantities for get and craft actions.
You should first reason step-by-step. This reasoning process MUST be enclosed within <thinking> </thinking> tags.
Then choose exactly one action and present it within <action> </action> tags.

[IF PRIVILEGED VIEW AND SUCCESSFUL TRAJECTORY]
A successful trajectory for the current task:

Guidance summary:
- Minimal plan: <...>
- Critical actions: <...>
- Checks: <...>
- Avoid: <...>

Use this information as a reference and continue solving the original task.

[ELSE IF PRIVILEGED VIEW AND FAILED TRAJECTORY]
A failed trajectory for the current task:

Failure analysis:
- Failure diagnosis: <...>
- Useful evidence: <...>
- Corrected plan: <...>
- Avoid: <...>

Use this information as a reference and continue solving the original task.
[END CONDITIONAL INSERT]
\end{promptlisting}

\clearpage
\begin{promptlisting}{TextCraft: later-step input}
[USER MESSAGE]
You are an expert agent operating in TextCraft. Your goal is: <GOAL_TEXT>
Prior to this step, you have already taken <STEP_COUNT> step(s). Recent history: [Observation 1: '<OBSERVATION_1>', Action 1: '<ACTION_1>'] [Observation 2: '<OBSERVATION_2>', Action 2: '<ACTION_2>'] ... [Observation K: '<OBSERVATION_K>', Action K: '<ACTION_K>']
[IF EPISODE-LEVEL CRAFTING COMMANDS ARE AVAILABLE]
Available crafting commands for this episode:
<CRAFTING_COMMANDS>
[END CONDITIONAL COMMAND CONTEXT]
Your current observation is:
<CURRENT_OBSERVATION>

Valid actions are: get N item, inventory, or craft N target item using N ingredient, N ingredient.
Use only the crafting commands listed above, in the observation, or already visible in the recent history.
You should first reason step-by-step within <thinking> </thinking> tags.
Then choose exactly one action within <action> </action> tags.

[IF PRIVILEGED VIEW AND SUCCESSFUL TRAJECTORY]
A successful trajectory for the current task:

Guidance summary:
- Minimal plan: <...>
- Critical actions: <...>
- Checks: <...>
- Avoid: <...>

Use this information as a reference and continue solving the original task.

[ELSE IF PRIVILEGED VIEW AND FAILED TRAJECTORY]
A failed trajectory for the current task:

Failure analysis:
- Failure diagnosis: <...>
- Useful evidence: <...>
- Corrected plan: <...>
- Avoid: <...>

Use this information as a reference and continue solving the original task.
[END CONDITIONAL INSERT]
\end{promptlisting}

\clearpage
\subsection{Text Experience Extraction}
Both tasks extract a summary from the task snapshot and the complete sequence
of executed actions and resulting observations, using the outcome-specific
prompts below. After removing the \texttt{<thinking>} block, accepted summaries
must have the matching outcome heading and four nonempty fields, without
extra headings or a copied trajectory. The accepted summary is cached for
the cycle.

\noindent\textbf{Evidence limits.} Each action and resulting observation is
capped at 240 and 480 characters, respectively. The task snapshot contains
at most the first four nonempty lines, each capped at 240 characters.

\begin{promptlisting}{Text summary: successful trajectory}
[USER MESSAGE]
You are writing concise guidance from one completed trajectory.
This guidance will be inserted into a future prompt for the same interactive task.

Output ONLY this format:
<thinking>
...
</thinking>

Guidance summary:
- Minimal plan: ...
- Critical actions: ...
- Checks: ...
- Avoid: ...

Rules:
- Use the complete trajectory as evidence, but do not copy the full task, observations, or command list.
- Use exact action strings in backticks when they are critical.
- Keep it concise and do not invent facts absent from the evidence.
- Think briefly inside <thinking> before writing the guidance summary.

Task snapshot:
<TASK_SNAPSHOT>

Successful trajectory evidence:
1. Action: `<ACTION_1>`
   Result: <OBSERVATION_AFTER_ACTION_1>
2. Action: `<ACTION_2>`
   Result: <OBSERVATION_AFTER_ACTION_2>
...
T. Action: `<ACTION_T>`
   Result: <OBSERVATION_AFTER_ACTION_T>
\end{promptlisting}

\clearpage
\begin{promptlisting}{Text summary: failed trajectory}
[USER MESSAGE]
You are writing concise guidance from one failed trajectory.
This guidance will be inserted into a future prompt for the same interactive task.

Output ONLY this format:
<thinking>
...
</thinking>

Failure analysis:
- Failure diagnosis: ...
- Useful evidence: ...
- Corrected plan: ...
- Avoid: ...

Rules:
- Use the complete failed trajectory as evidence, but do not copy the full task, observations, or command list.
- Distinguish useful partial progress from the action pattern that caused failure.
- Use exact action strings in backticks when they are critical.
- Keep it concise and do not invent facts absent from the evidence.
- Think briefly inside <thinking> before writing the failure analysis.

Task snapshot:
<TASK_SNAPSHOT>

Unsuccessful trajectory evidence:
1. Action: `<ACTION_1>`
   Result: <OBSERVATION_AFTER_ACTION_1>
2. Action: `<ACTION_2>`
   Result: <OBSERVATION_AFTER_ACTION_2>
...
T. Action: `<ACTION_T>`
   Result: <OBSERVATION_AFTER_ACTION_T>
\end{promptlisting}

\clearpage
\subsection{GUI Agent Prompt}
The complete call comprises a system message, the \texttt{mobile\_use} tool
definition, and a multimodal user message. Ordinary and privileged views
differ only in the inserted guideline. Both include all ground-truth prior
actions (\texttt{None} at the first step) and one current screenshot;
earlier screenshots and model predictions are excluded.

\begin{promptlisting}{GUI action input: system, tool, and user}
[SYSTEM MESSAGE]
You are a GUI agent. Use the mobile_use tool to complete the user's task from the current screenshot and prior action history.

# Response format

Response format for every step:
1) Thought: one concise sentence explaining the next move (no multi-step reasoning).
2) Action: a short imperative describing what to do in the UI.
3) A single <tool_call>...</tool_call> block containing only the JSON object.

Rules:
- Output exactly in the order: Thought, Action, <tool_call>.
- Be brief: one sentence for Thought, one for Action.
- Use normalized integer coordinates from 0 to 999.
- Do not output anything else outside those three parts.
- If finishing, use action=terminate in the tool call.

[TOOL DEFINITION]
{
  "type": "function",
  "function": {
    "name": "mobile_use",
    "description": "Use a touchscreen to interact with a mobile device. You can click, type, swipe, press system buttons, wait, and finish the task. Some applications take time to load, so wait when the UI has not settled. Coordinates are normalized integers from 0 to 999, where (0, 0) is the top-left corner; target the center of an element.",
    "parameters": {
      "type": "object",
      "properties": {
        "action": {
          "type": "string",
          "description": "Action to perform: click a point; long_press a point; swipe from coordinate to coordinate2; type into the active field; answer the user; press a system_button; open an app by name; wait; or terminate with status.",
          "enum": ["click", "long_press", "swipe", "type", "answer",
                   "system_button", "open", "wait", "terminate"]
        },
        "coordinate": {
          "type": "array", "items": {"type": "integer"},
          "description": "Required for click, long_press, and swipe: normalized [x,y]."
        },
        "coordinate2": {
          "type": "array", "items": {"type": "integer"},
          "description": "Required for swipe: normalized ending [x,y]."
        },
        "text": {
          "type": "string", "description": "Required for type, answer, and open."
        },
        "time": {
          "type": "number", "description": "Required for long_press and wait."
        },
        "button": {
          "type": "string", "enum": ["Back", "Home", "Menu", "Enter"],
          "description": "Required for system_button."
        },
        "status": {
          "type": "string", "enum": ["success", "failure"],
          "description": "Required for terminate."
        }
      },
      "required": ["action"]
    }
  }
}

[USER MESSAGE]
The user query: <USER_INSTRUCTION>

[IF PRIVILEGED VIEW]
Use this task guideline when choosing the next action:
<remark>
How to do:
1) <...>

Watch out:
- <...>
</remark>
[END CONDITIONAL INSERT]

Task progress (operations already completed on the current device):
[IF PREVIOUS ACTIONS EXIST]
Step 1: <GROUND_TRUTH_PREVIOUS_ACTION_1>
Step 2: <GROUND_TRUTH_PREVIOUS_ACTION_2>
...
Step t-1: <GROUND_TRUTH_PREVIOUS_ACTION_t-1>
[OTHERWISE]
None
[END CONDITIONAL HISTORY]
Choose the next action from the current screenshot.
<IMAGE: CURRENT_SCREENSHOT>
\end{promptlisting}

\clearpage
\subsection{GUI Experience Extraction}
The initial student independently attempts each logged step under ordinary
input; correctness follows AITZ EM (Appendix~\ref{app:gui_protocol}).
One accepted \texttt{<remark>} per trajectory is cached throughout training.
OEL and OEL + \MethodASelectionAbbr{} sample fresh ordinary-view responses at each
update (Appendix~\ref{app:optimization}).

\begin{promptlisting}{GUI summary input: system and complete user message}
[SYSTEM MESSAGE]
You are a conservative GUI error-correction summarizer. Correct-step actions and lines marked CORRECT action are the only ground truth. Lines marked WRONG attempt are prohibited negative evidence. Never turn a wrong attempt into a recommended or prerequisite operation, and never infer extra operations from the user query, screenshots, or task-world knowledge. If the demonstrated route appears incomplete or unusual, summarize only its shown correct actions. Keep the required output format exactly.

[USER MESSAGE]
Summarize this successful GUI trajectory into a short guideline for the student model.

User query: <USER_INSTRUCTION>

The student was tested independently at every step. Correct steps are text-only route context. A screenshot is shown only for a failed step, together with the student's attempted action and the correct action. Focus on the failed steps and the difference between the attempt and the correct action.

For failed steps, "WRONG attempt — DO NOT DO" is negative evidence shown only to diagnose the mistake. "CORRECT action — DO THIS" is the authority. Build the route only from correct-step actions and CORRECT actions. A wrong attempt may appear only as an explicit "do not" warning followed by the correct alternative.

[FOR EACH STEP, IN TRAJECTORY ORDER]
[IF THE STUDENT ACTION IS CORRECT]
Step <INDEX> — student handled correctly
Action: <CANONICAL_CORRECT_ACTION>
[OTHERWISE]
Step <INDEX> — student failed
<IMAGE: SCREENSHOT_AT_THIS_STEP>

CORRECT action — DO THIS: <CANONICAL_CORRECT_ACTION>
WRONG attempt — DO NOT DO: <STUDENT_ATTEMPT>
[END STEP BRANCH; REPEAT FOR ALL STEPS]

Briefly cover the demonstrated route only; do not invent intermediate actions. Spend most detail on failed steps. Preserve text that must be typed, distinguish taps, swipes, system buttons, waiting, and completion, and state what visible UI cue should trigger each uncertain action. Include one useful fallback. Before answering, verify that every failed step's CORRECT action is represented and no WRONG attempt is written as a positive instruction. Do not include coordinates or tool calls.

Output only:
<remark>
How to do:
1) ...

Watch out:
- ...
</remark>
\end{promptlisting}
\par\clearpage\endgroup